\documentclass{article}

\usepackage{arxiv}

\usepackage[utf8]{inputenc} 
\usepackage[T1]{fontenc}    
\usepackage{hyperref}       
\usepackage{url}            
\usepackage{booktabs}       
\usepackage{amsfonts}       
\usepackage{amsmath}        
\usepackage{amssymb}
\usepackage{nicefrac}       
\usepackage{microtype}      
\usepackage{lipsum}		
\usepackage{graphicx}
\usepackage{natbib}
\usepackage{doi}
\usepackage{array, tabularx, caption, boldline}
\usepackage{tikz}
 \usepackage[cal=boondoxo]{mathalfa}
\usepackage{float}
\usepackage{comment}
\usepackage{color}
\usepackage{pgfplotstable}
\usepackage{pgfplots}
\usepackage{booktabs}
\usepackage{textcomp}
\usepackage{xcolor}
\usepackage{colortbl}
\graphicspath{{Figures/}}     
\pgfplotsset{width=10cm,compat=1.9}
\usetikzlibrary{shapes,arrows}
\usetikzlibrary{positioning}

\title{LWGNet: Learned Wirtinger Gradients for Fourier Ptychographic Phase Retrieval}

\author{{Atreyee Saha}, {Salman S Khan}, {Sagar Sehrawat}, {Sanjana S Prabhu}, {Shanti Bhattacharya}, {Kaushik Mitra} \\
	Department of Electrical Engineering \\
	Indian Institute of Technology Madras\\
	Chennai, India 600036 \\
}

\renewcommand{\headeright}{ }
\renewcommand{\undertitle}{ }
\renewcommand{\shorttitle}{LWGNet}

\hypersetup{
pdftitle={LWGNet},
pdfsubject={cs},
pdfauthor={Atreyee Saha},
pdfkeywords={Fourier ptychography, Phase imaging},
}

\begin{document}
\maketitle

\begin{abstract}
Fourier Ptychographic Microscopy (FPM) is an imaging procedure that overcomes the traditional limit on Space-Bandwidth Product (SBP) of conventional microscopes through computational means. It utilizes multiple images captured using a low numerical aperture (NA) objective and enables high-resolution phase imaging through frequency domain stitching. Existing FPM reconstruction methods can be broadly categorized into two approaches: iterative optimization based methods, which are based on the physics of the forward imaging model, and data-driven methods which commonly employ a feed-forward deep learning framework. We propose a hybrid model-driven residual network that combines the knowledge of the forward imaging system with a deep data-driven network. Our proposed architecture, LWGNet, unrolls traditional Wirtinger flow optimization algorithm into a novel neural network design that enhances the gradient images through complex convolutional blocks. Unlike other conventional unrolling techniques, LWGNet uses fewer stages while performing at par or even better than existing traditional and deep learning techniques, particularly, for low-cost and low dynamic range CMOS sensors. This improvement in performance for low-bit depth and low-cost sensors has the potential to bring down the cost of FPM imaging setup significantly. Finally, we show consistently improved performance on our collected real data.\footnote{We have made the code avaiable at: \hyperref[https://github.com/at3e/LWGNet.git]{https://github.com/at3e/LWGNet.git}}.
\end{abstract}


\section{Introduction}
\label{sec: Intro}
One of the main challenges in medical computational imaging is to make imaging technology accessible for point of care diagnostics in resource constrained communities. One way to make these techniques accessible includes designing cheaper and portable hardware. \cite{breslauer2009mobile,smith2011cell} have used smartphone cameras for microscopy, while \cite{aidukas2019low} have used 3D printed microscopes for wide field of view high resolution imaging. However, the use of cheaper and inefficient hardware introduces limits on the imaging capabilities of these systems through various degradations which can only be dealt by designing effective computational algorithms.

In this work, we will be focusing on a particular microscopy technique called Fourier Ptychographic Microscopy (FPM) \cite{zheng2013wide}. It is a computational microscopy technique that allows us to perform high resolution and wide field of view (FOV) imaging. It circumvents the limit on SBP of conventional microscope by relying on multiple low-resolution captures of the sample under programmed illumination and reconstruction via phase retrieval algorithms. Development in FPM in the recent years has included improving the temporal resolution through multiplexed illumination \cite{Tian:14}, designing better reconstruction algorithms \cite{boominathan2018phase,Bian:15,bostan2018accelerated}, and designing better illumination codes \cite{kellman2019data}.

Despite the above-mentioned progress, there hasn’t been enough work done to make these FPM systems accessible for point of care diagnostics. The following fundamental challenges have been the main reason behind that. First, the existing FPM systems, like most low-light imaging modalities, suffer from significant noise and dynamic range problems especially for darkfield images. These FPM systems rely on expensive optics and scientific grade sCMOS sensors to increase the SNR and dynamic range of the low-resolution darkfield captures which increases the system cost significantly \cite{Tian:14}. Second, existing reconstruction algorithms designed for FPM systems have shown results only for these expensive systems and are unlikely to perform optimally when the quality of sensor degrades. Third, the existing reconstruction techniques are either slow or are model-independent data-driven techniques that ignore the forward imaging process completely. One related work on making FPM systems accessible is \cite{aidukas2019low}. However, the authors rely on extensive pre-processing and calibration prior to reconstructing with traditional algorithms. As a result, their method is not end to end and doesn't exploit the advantage of data-driven techniques.

Keeping the above challenges in mind, we propose `LWGNet' for sequential FPM reconstruction. LWGNet is a novel physics-driven unrolled neural network that combines the expressiveness of data-driven techniques with the interpretability of iterative Wirtinger flow based techniques \cite{Bian:15}. Unlike existing unrolled networks, LWGNet performs most of its operation on image gradient through \textit{complex-valued} operations. Specifically, it learns a non-linear mapping from complex stochastic gradients to intermediate object fields through complex-valued neural networks. Such a learned mapping helps preserve the high frequency details in the peripheral darkfield images, especially for low dynamic range sensor. LWGNet outperforms both traditional and deep-learning methods in terms of reconstruction quality, especially for low-cost machine vision sensors with poor dynamic range. We show this by performing extensive evaluations on simulated and real histopathological data captured under different bit depths. 

In summary, we make the following contributions:
\begin{itemize}
    \item We propose LWGNet which is a physics-inspired complex valued feed-forward network for FPM reconstructions that exploits the physics of the FPM model and data-driven methods.
    \item The proposed approach uses a learned complex-valued non-linear mapping from gradients to object field that helps restore the finer details under a low dynamic range scenario thereby reducing the gap in reconstruction quality between expensive HDR sCMOS cameras and low cost 8-bit CMOS sensors. This enables reducing the cost of the experimental setup to a large extent.
    \item We collect a real dataset of 8, 12 and 16 bit low resolution measurements using a CMOS sensor along with the corresponding aligned groundtruth for finetuning our method. To the best of our knowledge, this is the only FPM dataset captured with multiple bit depth settings using a low cost CMOS sensor. This dataset will be made public upon acceptance.
    \item The proposed network outperforms existing traditional and learning based algorithms in terms of reconstruction quality for both simulated and real data as verified through extensive experiments on challenging histopathological samples.
    
\end{itemize}
\begin{figure}[H]
    \centering
    \includegraphics[width=0.8\textwidth]{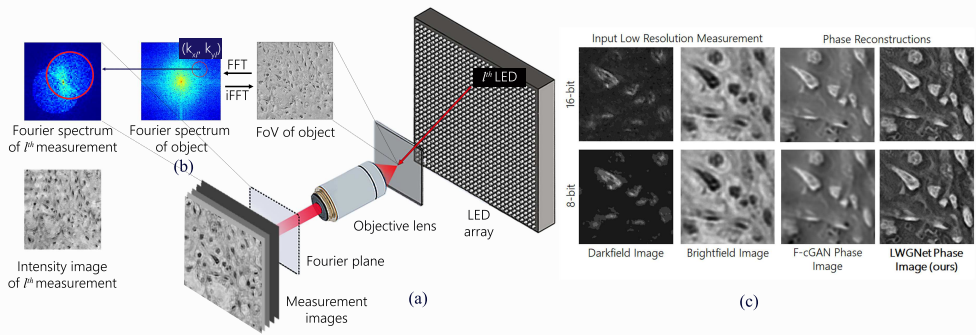}
    \caption{Overview of FPM phase reconstruction. (a) The object is placed between a planar LED array and objective of the inverted microscope. Low resolution images corresponding to $\mathcal{l}^{\text{th}}$ LED is captured. (b) Representation of the equivalent forward model, comprises of down-sampling the object spectrum by the system pupil function centred at $(k_{x\mathcal{l}}, k_{y\mathcal{l}})$. (c) Examples of input low resolution measurements under different sensor quantization and corresponding reconstructions using F-cGAN and the proposed LWGNet.}
    \label{overview_of_model}
\end{figure}

\section{Background}
\label{sec:background}
\subsection{The nature of phase retrieval problem}
\label{sec: ph_eqns}
We are interested in the case of phase recovery from intensity measurements only. The experimental setup for FPM involves illuminating the object from multiple angles and is described in detail in \cite{zheng2013wide}. Mathematically, the phase retrieval problem while using multiple structured illuminations for intensity measurement is as follows:
\begin{align}
    \text{Find} \hspace{0.5em} \mathcal{O} , \hspace{1em}
    \text{subject to} \hspace{1em} I_{\mathcal{l}} = \vert \mathcal{F}^{-1} \{ P_{\mathcal{l}} \odot \mathcal{F} \{ \mathcal{O}(\textbf{k}-\textbf{k}_{\mathcal{l}}) \} \} \vert ^2 , \text{for} \hspace{0.5em} \mathcal{l} = 1, 2, \hdots , L ,
\end{align}
where $\mathcal{F}$ is the 2D spatial Fourier transform and $\mathcal{F}^{-1}$ is the corresponding inverse transform. $ P_{\mathcal{l}} \in \mathbb{C}^{N\times N} $ is the pupil function for $\mathbf{k}_{\mathcal{l}} = (k_{x\mathcal{l}}, k_{y\mathcal{l}})$, the unique spatial frequency corresponding to the $\mathcal{l}^{\text{th}}$ illumination source.
It is assumed that $\mathcal{O}$ is an optically thin sample, i.e. its transmittance is close to unity. Classical retrieval algorithms seek to solve the following minimization problem,
\begin{equation}
    \underset{\mathcal{O}}{\text{min}} \sum_{\mathcal{l}=1}^L \Vert I_{\mathcal{l}} - \lvert A_{\mathcal{l}} \{ \mathcal{O} \} \rvert^2 \Vert_2^2 ,
   \label{eq: obj_fun}
\end{equation}
where $A \{ . \} \triangleq \mathcal{F}^{-1} \circ P_{\mathcal{l}} \odot \mathcal{F} \{ . \}$.

The objective is a non-convex, real-valued function of complex-valued object field. Conventional approaches like gradient descent will converge to a stationary point. 

\subsection{Related Works}
\label{sec: lit_survey}
There have been numerous works on FPM reconstruction. Ou et al.\cite{Ou:13} successfully performed whole slide high-resolution phase and amplitude imaging using a first-order technique based on alternate projections (AP). Song et al.\cite{song2019full} proposed algorithms to overcome system aberration in FPM reconstruction. In \cite{bian2016motion}, the authors proposed sample motion correction for dynamic scenes. The work by Tian et al. \cite{Tian:14} extended the AP algorithm for multiplexed illumination in FPM setup. Besides AP, Wirtinger flow based methods are shown to perform well under low SNR \cite{Bian:15}. The Wirtinger flow algorithm is also extended to the multiplexed scheme by Bostan et al. \cite{bostan2018accelerated}. 

Recently, deep learning techniques have also been explored to solve the FPM reconsruction problem. Jiang et al. \cite{Jiang:18} show a novel approach by treating the object as a learnable network parameter. However, as it requires optimization for each patch, it has a large inference time. Kappeler et al. \cite{kappeler2017ptychnet} have performed high-resolution amplitude recovery using a CNN architecture. Nguyen et al. \cite{Nguyen:18} have performed time-lapse high-resolution phase imaging via adversarial training of a U-Net architecture \cite{mirza2014conditional}. The authors performed phase imaging using a limited set of low-resolution images.
The authors train (transfer learn in the case of new biological type of cells) U-Net on a subset of low resolution video measurements. While testing, they predict phase of dynamic cell samples on the subsequent frames. FPM phase reconstruction under various overlap conditions between adjacent low-resolution images in the frequency domain has been studied by Boominathan et al. \cite{boominathan2018phase}. However, they show results only on simulated samples. Kellman et al.\cite{kellman2019data} have designed a Physics-based Neural Network (PbNN) that learns patterns for multiplexed illumination by optimizing the weights of the LEDs.

An important line of work is the employment of high-resolution phase imaging using low-cost components. To our best knowledge, the only attempt made so far is by Aidukas et al. \cite{aidukas2019low}, who employed traditional reconstruction algorithm accompanied with an elaborate calibration and pre-processing of low-resolution images taken from a commercial-grade camera. The authors performed amplitude reconstruction, but their system has fundamental limitations due to sensor size and optical aberrations.

\section{Proposed Method: LWGNet}
\label{sec:method}
\begin{figure}
    \centering
    \includegraphics[width=0.85\textwidth]{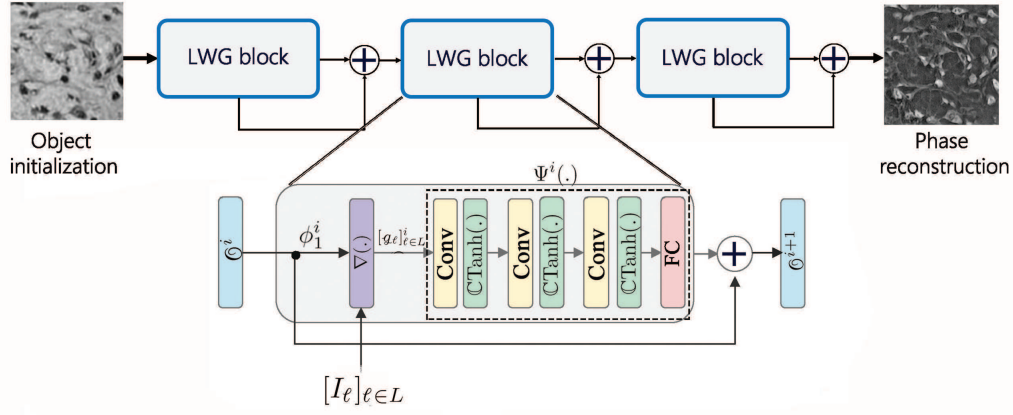}
    \caption{Model architecture of LWGNet comprising of three stages of the LWG update block.The inputs are object and the pupil initialization and measurement images for every stage. LWG block comprises of a Gradient computation block followed by three stages of Conv+$\mathbb{C}$Tanh layers followed by a Fully Connected(FC) layer. The adder block updates the input object field using the modified gradients.}
    \label{fig:proposed_method}
\end{figure}
The naive Wirtinger flow algorithm seeks to minimize the following objective\cite{Bian:15},

\begin{align}\label{eq:objective}
   f(\mathcal{O}) &=  \sum_{\mathcal{l}=1}^L \Vert I_{\mathcal{l}} - \lvert A_{\mathcal{l}} \{\mathcal{O}\} \rvert^2 \Vert_2^2 \\
   &= \sum_{\mathcal{l}=1}^L \Vert I_{\mathcal{l}} - A_{\mathcal{l}} \{\mathcal{O}\} \odot \overline{A_{\mathcal{l}} \{\mathcal{O}^i\}} \Vert_2^2,
   \label{eq: obj_fun_Wirtinger}
\end{align}
where $A\{ . \} \triangleq \mathcal{F}^{-1} \circ P_\mathcal{l} \odot \mathcal{F} \{ . \}$, $\mathcal{F}$ is the 2D spatial Fourier transform, $\mathcal{F}^{-1}$ is the corresponding inverse transform, $P_{\mathcal{l}}$ is the pupil function corresponding to the $\mathcal{l}^{\text{th}}$ illumination source and $I_\mathcal{l}$ is the capture $\mathcal{l}^{\text{th}}$ measurement for the same illumination source. Minimizing Eq.(\ref{eq: obj_fun_Wirtinger}) with respect to $\mathcal{O}$ using gradient descent leads to the following  $i^{\text{th}}$ step update, 
\begin{align}
    \mathcal{O}^{i+1} = \mathcal{O}^i - \eta \frac{1}{L} \sum_{\mathcal{l}=1}^{L}\nabla_{\mathcal{O}}f^i_{\mathcal{l}},
\label{eq: gradient_update}
\end{align}
where $\eta$ is the step-size and, 
\begin{align}
    \nabla_{\mathcal{O}}f^i_{\mathcal{l}} =  A_{\mathcal{l}}^{H} \{ ( \lvert A_{\mathcal{l}} \{\mathcal{O}^i\} \rvert^2 - I_{\mathcal{l}}) \odot A_{\mathcal{l}} \{\mathcal{O}^i\} \}.
    \label{eq:grad_Wirtinger}
\end{align}
A naive way to unroll the above gradient descent Wirtinger flow algorithm could be to design a neural network with each stage performing the gradient update step of Eq.(\ref{eq: gradient_update}) followed by convolutional block. However, we found that such a naive unrolling doesn't converge to any meaningful solution. To overcome this, we design a K-stage unrolled network with the $i^{\text{th}}$ stage performing the following operation 
\begin{align}
    \mathcal{O}^{i+1} = \mathcal{O}^i + \Psi^i([\mathcal{g}_{\mathcal{l}}]_{\mathcal{l}\in L}^i).\label{eq:res_block}
\end{align}
Here,  
\begin{align}
   \mathcal{g}_{\mathcal{l}}^i &= A_{\mathcal{l}}^{H} \{ ( \lvert A_{\mathcal{l}} \{\phi^i_{\mathcal{l}-1}\} \rvert^2 - I_{\mathcal{l}}) \odot A_{\mathcal{l}} \{\phi^i_{\mathcal{l}-1}\} \},\label{eq:stoch_grad}
\end{align}
\begin{align}
    \phi^i_{\mathcal{l}} &= \phi^i_{\mathcal{l}-1}- \eta\mathcal{g}_{\mathcal{l-1}}^i, 
\end{align}
$\phi^i_1 = \mathcal{O}^i$, $\phi_1^0 = \mathcal{O}^0= I_0$, $[\mathcal{g}_{\mathcal{l}}]_{\mathcal{l}\in L}^i$ is a stack of $\mathcal{g}_{\mathcal{l}}^i$ and $\Psi(.)$ is a learned complex neural network. $\mathcal{g}_{\mathcal{l}}^i$ in the above equations can be interpreted as stochastic gradients corresponding to each illumination source. $\Psi(.)$ then learns to non-linearly combine these stochastic gradients and update the object field as shown in Eq.(\ref{eq:res_block}). Experimentally, we found that such an unrolled network converged faster with fewer K than naively unrolled wirtinger flow. $\Psi(.)$ consists of 3 $3\times 3$ complex convolutions followed by complex Tanh non-linearity, instance norm,  and a fully connected layer. The learned non-linear $\Psi(.)$ helps combine the stochastic gradients in a more effective way especially for peripheral darkfield images which are typically degraded due to lower bit depth and noise.

We simulate complex arithmetic using real-valued entities \cite{trabelsi2018deep} in $\Psi(.)$. We perform equivalent 2D complex-valued convolution using real-valued CNN blocks as follows: let there be $M$ input channels and $N$ output channels for a given CNN layer. Then, complex-valued filter matrix weight for $m^{\text{th}}$ input channel and $n^{\text{th}}$ output channel is $\textbf{w}_{mn} = \textbf{a}_{mn} + j\textbf{b}_{mn}$ that convolves with a 2D complex input $\text{z}_m = \text{x}_m +j \text{y}_m$. Here, $\textbf{a}_{mn}$  and  $\textbf{b}_{mn}$ are real-valued kernels, and $\text{x}_m$ and  $\text{y}_m$ are also real.
\begin{align}\label{eq:conv}
     \textbf{w}_{mn} \ast \text{z}_m = \textbf{a}_{mn} \ast \text{x}_m + j \textbf{b}_{mn} \ast \text{y}_m.
\end{align}
We initialize filter weights as a uniform distribution. The CNN layer is followed by an 2D instance normalization layer. An amplitude-phase type non-linear activation function acts on the normalised outputs  given by: 
\begin{align}
    \mathbb{C}\text{Tanh}(\text{z}_k) = \text{Tanh}(|\text{z}_k|)e^{i\theta_{\text{z}_k}}.
\end{align}
Finally, after 3 convolutional and non-linearity blocks, we use a fully connected block that acts on the channel dimension.

\subsection{Loss Functions}
\label{sec:loss_func}
Let $\mathcal{O}$ be the ground truth object and $\hat{\mathcal{O}}$ be the reconstruction. We use the following weighted combination of loss functions to optimize our network,
\begin{align}\label{eq:loss_total}
    \mathcal{L} = \lambda_1 \mathcal{L}_{MSE} + \lambda_2 \mathcal{L}_{FMAE} + \lambda_3 \mathcal{L}_{VGG} 
\end{align}
where,
\begin{align}\label{eq:loss_mse}
    \mathcal{L}_{MSE} = \Vert \angle(\hat{\mathcal{O}}) - \angle(\mathcal{O}) \Vert_2^2 +  \Vert |\hat{\mathcal{O}}| - |\mathcal{O}| \Vert_2^2 
\end{align}    
\begin{align}\label{eq:loss_fmae}
    \mathcal{L}_{FMAE} =  \Vert |\mathcal{F}(\hat{\mathcal{O}})| - |\mathcal{F}(\mathcal{O})|\Vert_1 
\end{align}    
\begin{align}\label{eq:loss_vgg}
    \mathcal{L}_{VGG/i,j} = \beta_1 \Vert \psi_{i,j}(|\hat{\mathcal{O}}|) - \psi_{i,j}(|\mathcal{O}|) \Vert_2^2 + \beta_2 \Vert \psi_{i,j}(\angle\hat{\mathcal{O}}) - \psi_{i,j}(\angle\mathcal{O}) \Vert_2^2
\end{align}
$\Vert \cdot \Vert_p$ represents the $p$-norm. Eq.(\ref{eq:loss_mse}) defines the pixel-wise Mean Squared Error (MSE) loss over amplitude and phase components. The VGG loss function is a perceptual loss function, as defined by \cite{ledig2017photorealistic}. It minimizes the MSE from the output  of the feature maps of the pre-trained VGG-19 network. $\psi_{i,j}(.)$ is  the  feature  map  obtained  by  the  $j$-th  convolution (after activation) before the $i$-th max-pooling layer in the network, which we consider given. Here, we use $\text{ReLU}$ output $\psi_{2,2}$ and $\psi_{4,3}$ the VGG-19 network.  The output feature maps corresponding to the reconstruction  and ground truth amplitudes are compared. Eq.(\ref{eq:loss_fmae}) minimizes the $L_1$ -norm of the magnitude Fourier spectrum between reconstruction and ground truth.

\section{Experiments and Results}\label{sec:expt}
\subsection{Simulated Dataset}\label{dataset_sec}
Images in Iowa histology dataset \cite{Iowa} are used for simulating objects fields with uncorrelated amplitude and phase. The FoV of the entire histology slide is divided into $320\times320$ pixels; the amplitude images are normalised, such that the values lie in the range $[0,1]$. The pixel values of the phase image are linearly mapped to the range $[-\pi, \pi]$. These objects are further divided into training, validation and test splits. Then the FPM forward model \cite{zheng2013wide}, is used to generates low resolution intensity images of $64\times64$ pixels from these object field samples.
\subsection{Our Captured Real Dataset}
\label{sec:real_dataset}
The experimental setup to capture real data consists of a Nikon Eclipse TE300 inverted microscope and a 10X/0.25 NA objective lens for imaging. An AdaFruit programmable RGB LED array  ($32\times32$, planar) is used to illuminate the sample from various angles using Red LED with a wavelength of 630nm. By using central 225 LEDs, we captured sequential low-resolution images using a High Dynamic Range (HDR) scientific grade sCMOS camera (PCO Edge 5.5) and a low cost machine vision CMOS camera (FLIR FLEA3). The sCMOS camera allows us to capture only 16-bit images while the CMOS allows to capture 8, 12 and 16-bit images. The sCMOS camera has a resolution of $2560\times2160$ pixels and a pixel pitch $6.5\mu m$ while the CMOS sensor has a resolution of $1552\times2080$ pixels and pixel pitch of $2.5\mu m$. Both the sensors are monochromatic. The sCMOS camera is used on the front port of the inverted microscope while the CMOS camera is used on the side port of the same microscope.

The LEDs on the array have a grid spacing of 4mm and the array is placed at a distance of 80 mm from the sample plane of the microscope. The illumination from the LED array is controlled using an Arduino MEGA 2560 microcontroller, and simultaneously sends a clock signal to external exposure start port of the sCMOS camera for controlling the camera shutter speed with respect to LED illumination.

Before testing our network on real data, we finetune our network on a small set of real training data. To capture a training set, first low resolution images are captured in a sequential manner using the sCMOS sensor and the CMOS sensor. To generate the groundtruth data for training, FPM phase reconstructions using the algorithm in \cite{Tian:15} is used on the captured low resolution sCMOS images. To compensate for the misalignment between sCMOS and CMOS images, image registration is also performed. In total 8 slides of cervical, cerebral cortical, lung carcinoma and osteosarcoma cells were used for real data. 4 slides were kept for training and and 4 for testing. We show the capture setup and few real captured data in the supplementary.

We implement our neural network using PyTorch \cite{NEURIPS2019_9015}, and train the same on 2 GTX 1080Ti GPUs of 12GiB capacity, for 100 epochs. The loss function parameters as described in Eq.(\ref{sec:loss_func}) are $(\lambda_1, \lambda_2, \lambda_3) = (0.1, 0.05, 1)$ and $(\beta_1, \beta_2) = (0.5, 1)$. We use Adam optimizer with an initial learning rate (LR) of $10^{-4}$ and a learning rate scheduler that reduces the LR by a factor of 0.1 when the overall loss does not improve. On the simulated data, we use 3 stages of the update block as described in Fig \ref{fig:proposed_method} for 16-bit and 12-bit depth images, and increase the number of stages to 5 for 8-bit images. 

\subsection{Comparison on Simulated Data}\label{sec:sim_results}
\begin{table}[!ht]
    \centering
     \begin{tabular*}{0.8825\linewidth}{|c|c|}
    \hline
         \hbox{\rotatebox{90}{\hspace{2em}16-bit reconstruction}} &  \fcolorbox{white}{white}{\includegraphics[width=0.8\textwidth]{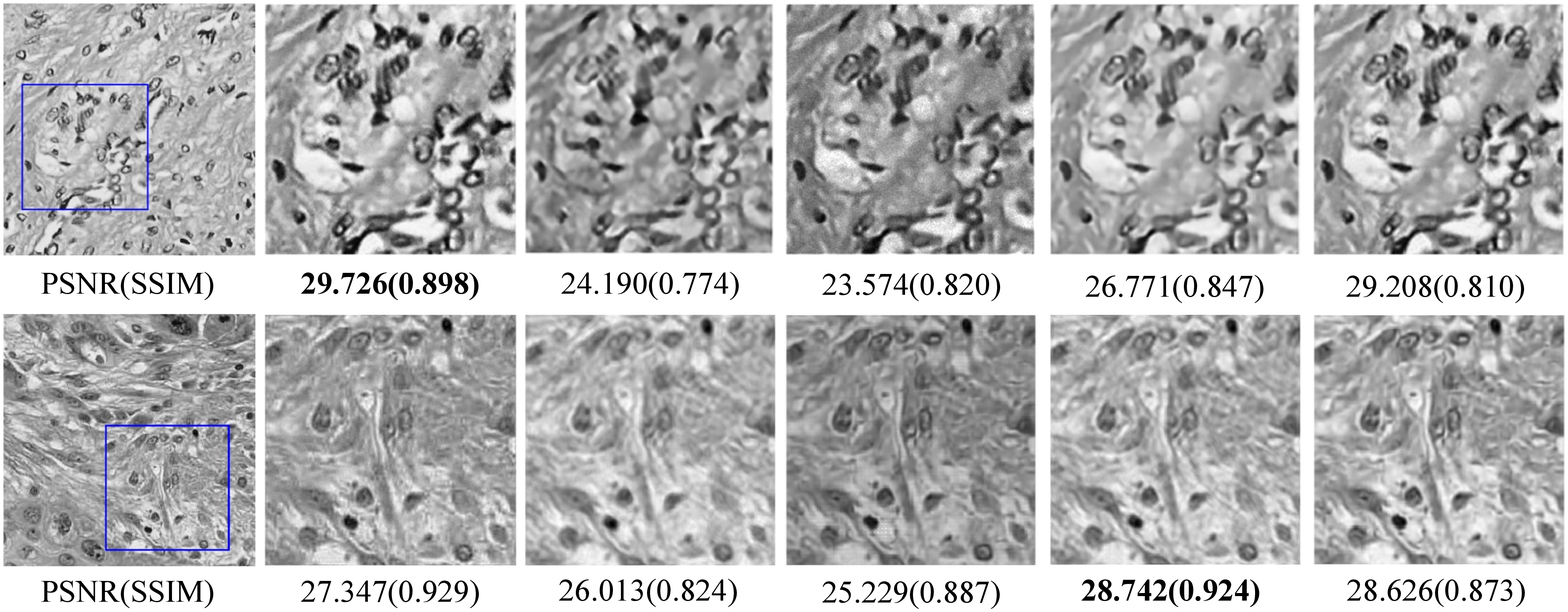}}\\
    \hline
         \hbox{\rotatebox{90}{\hspace{2em}12-bit reconstruction}} &  \fcolorbox{white}{white}{\includegraphics[width=0.8\textwidth]{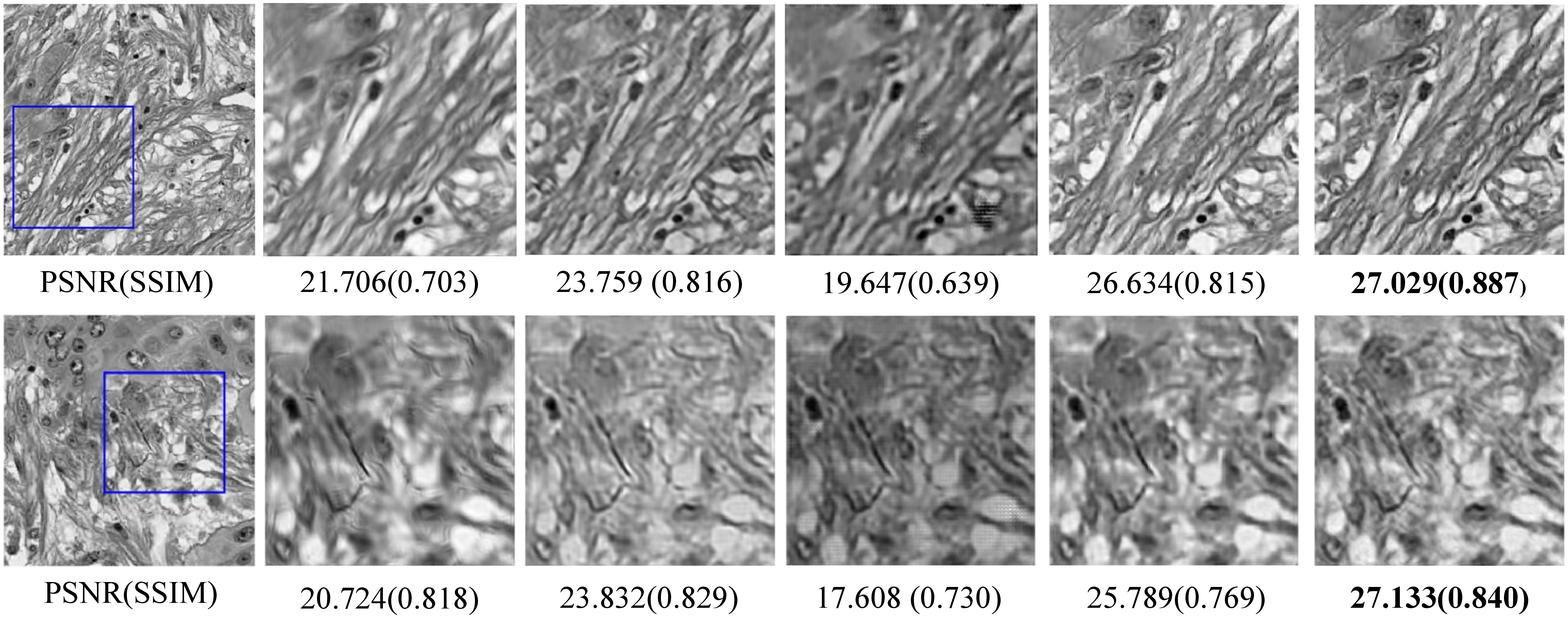}}\\
    \hline
         \hbox{\rotatebox{90}{\hspace{3em}8-bit reconstruction}} &  \fcolorbox{white}{white}{\includegraphics[width=0.8\textwidth]{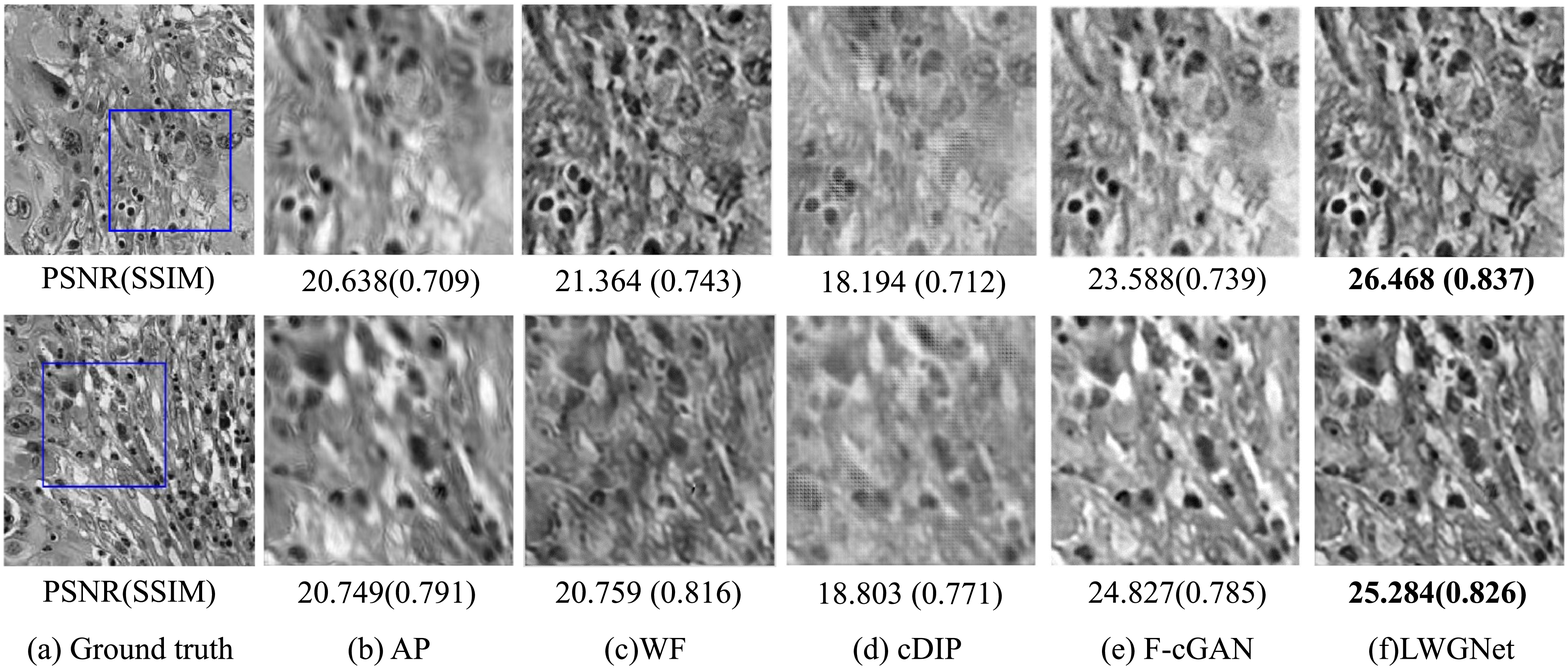}} \\
         
    \hline
    
    \end{tabular*}
    \vspace{2em}
    \captionof{figure}{Phase reconstruction on simulated data using proposed LWGNet method are perceptually superior at lower-bit depth depths compared to existing methods. Traditional algorithms WF and AP do not perform well under lower bit-depths. Among deep methods, cDIP performs poorly at lower bit depths.}
    \label{fig:sim_results}
\end{table}

\begin{table}[!ht]
\centering
\pgfplotstabletypeset[
column type=c,
every head row/.style={
before row={
\toprule
Method & \multicolumn{2}{c}{16} & \multicolumn{2}{c}{12} & \multicolumn{2}{c}{8} & Memory & Inference time & Parameters\\
\cline{2-7}
},
after row=\midrule,
},
every even row/.style={
before row={\rowcolor[gray]{0.9}}
},
every last row/.style={
after row=\bottomrule},
columns/method/.style ={column name=},
columns/psnr16/.style ={column name=PSNR},
columns/ssim16/.style={column name=SSIM},
columns/psnr12/.style ={column name=PSNR},
columns/ssim12/.style={column name=SSIM},
columns/psnr8/.style={column name=PSNR},
columns/ssim8/.style={column name=SSIM},
columns/memory/.style={column name= \strut{\textit{(in GiB)}}},
columns/time/.style={column name= \strut{\textit{(in ms)}}},
columns/param/.style={column name= \strut{\textit{(in millions)}}},
col sep=&, row sep=\\,
string type,
]{
method & psnr16 & ssim16 & psnr12 & ssim12 & psnr8 & ssim8 & memory & time & param\\
AP & 24.346 & 0.835 & 21.764 & 0.728 & 20.500 & 0.594 & NA & NA & NA\\
WF & 23.172 & 0.769 & 22.706 & 0.744 & 21.621 & 0.730 & NA & NA & NA\\
cDIP & 25.654 & 0.815 & 22.563 & 0.718 & 20.951 & 0.712 & 3.24 & 5& 54.9\\
F-cGAN & \textbf{29.021} & \textbf{0.907} & \underline{26.325} & \underline{0.797} & \underline{25.715} & \underline{0.765} & 3.52 & 29 & 7.88\\
\textbf{LWGNet} & \underline{28.745} & \underline{0.829} & \textbf{27.726} & \textbf{0.807} & \textbf{26.035} & \textbf{0.802} & 2.26 & 341 & 0.39\\
}
\vspace{2em}
\caption{Compared to previous works, the proposed LWGNet achieves a better reconstruction quality at lower bit-depths using fewer trainable parameters.}
\label{tab:sim_metric}
\end{table}

We compare our simulated and experimental results against existing iterative procedure and deep-learning techniques. Under iterative methods, we have compared with the alternate projections (AP) algorithm proposed by Tian et al. \cite{Tian:14}. Another class of iterative method for phase retrieval is the  Wirtinger Flow Optimization proposed by \cite{Bian:15}. Under data-driven techniques, we consider the conditional deep image prior (cDIP) architecture proposed by Boominathan et al. \cite{boominathan2018phase}. We also compare against Fourier-loss cGAN (F-cGAN) architecture provided by Nguyen et al. \cite{Nguyen:18} and modify the input layer dimension of the U-Net architecture to take into account the difference in number of illumination LEDs. 

Table \ref{tab:sim_metric} shows image quality with two metrics, namely Peak Signal to Noise Ratio (PSNR) and Structural SIMilarity (SSIM) against existing FPM phase retrieval algorithms. The proposed algorithm is shown to perform at par for the 16-bit simulation results and outperforms the baseline methods at lower bit depths of 12-bit and 8-bit in terms of both PSNR and SSIM. Additionally, the proposed model involves fewer trainable parameters and requires about 100 epochs of training. The proposed method is shown to be less memory intensive compared to the deep-learning based models, cDIP and F-cGAN. However, the Gradient computation block is computationally intensive taking longer inference time, compared to matrix computations. The memory and inference time reported in Table \ref{tab:sim_metric} are for a 3 stage proposed network. Fig \ref{fig:sim_results} shows a few phase reconstructions from simulated data at three bit depths. For 16-bit measurement images, both traditional and neural network based methods perform competitively well. Fig \ref{fig:sim_results}(b) shows AP-based reconstruction method \cite{Tian:14} suffers the most from degradation of peripheral LED measurements at lower bit-depths. Gradient-based methods \cite{Bian:15} display poor contrast and blurring at the edges as shown in Fig \ref{fig:sim_results}(c). Consequently, this is corrected for in the proposed method as shown in \ref{fig:sim_results}(f). cDIP \cite{boominathan2018phase} shows checkerboard artifacts that become more prominent with under lower bit depth. Similarly, F-cGAN \cite{Nguyen:18} model successfully preserves finer details at higher bit-depths, but shows blur artifacts with increase in quantization.

\subsection{Comparison on Real Data}
\begin{table}[!ht]
    \centering
     \begin{tabular*}{0.9325\linewidth}{|c|c|}
    \hline
         \hbox{\rotatebox{90}{\hspace{0em}16-bit reconstruction}} &  \fcolorbox{white}{white}{\includegraphics[width=0.85\textwidth]{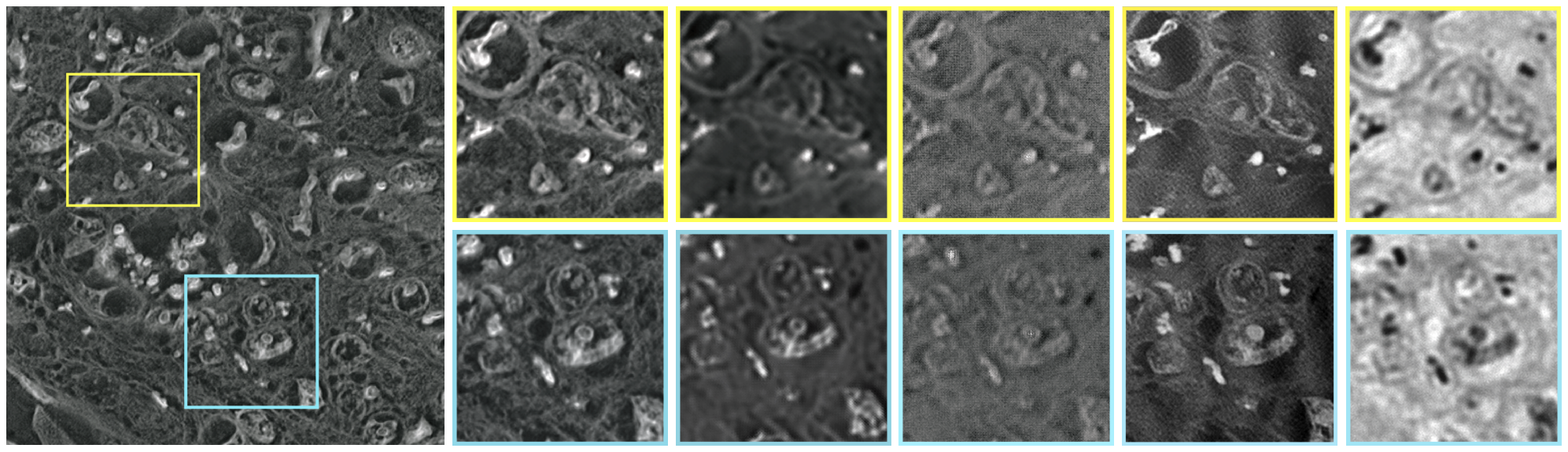}}\\
    \hline
         \hbox{\rotatebox{90}{\hspace{0em}16-bit reconstruction}} &  \fcolorbox{white}{white}{\includegraphics[width=0.85\textwidth]{Figures/Captured/EXP-16B-R1.eps}}\\
    \hline
         \hbox{\rotatebox{90}{\hspace{0em}12-bit reconstruction}} &  \fcolorbox{white}{white}{\includegraphics[width=0.85\textwidth]{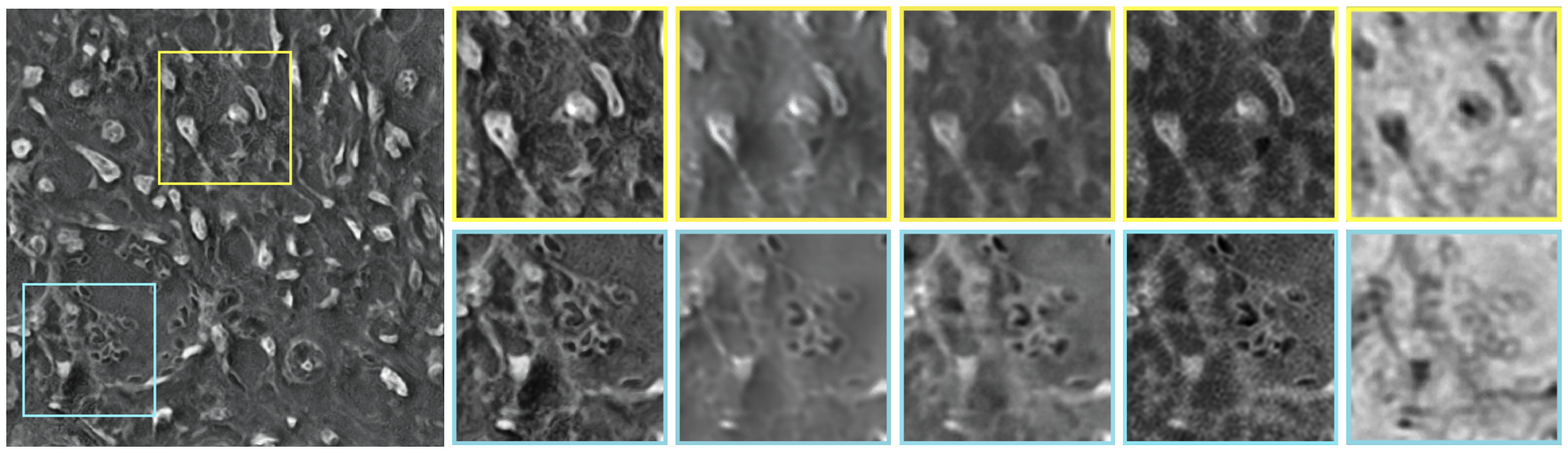}}\\
    \hline
         \hbox{\rotatebox{90}{\hspace{1em}8-bit reconstruction}} &  \fcolorbox{white}{white}{\includegraphics[width=0.85\textwidth]{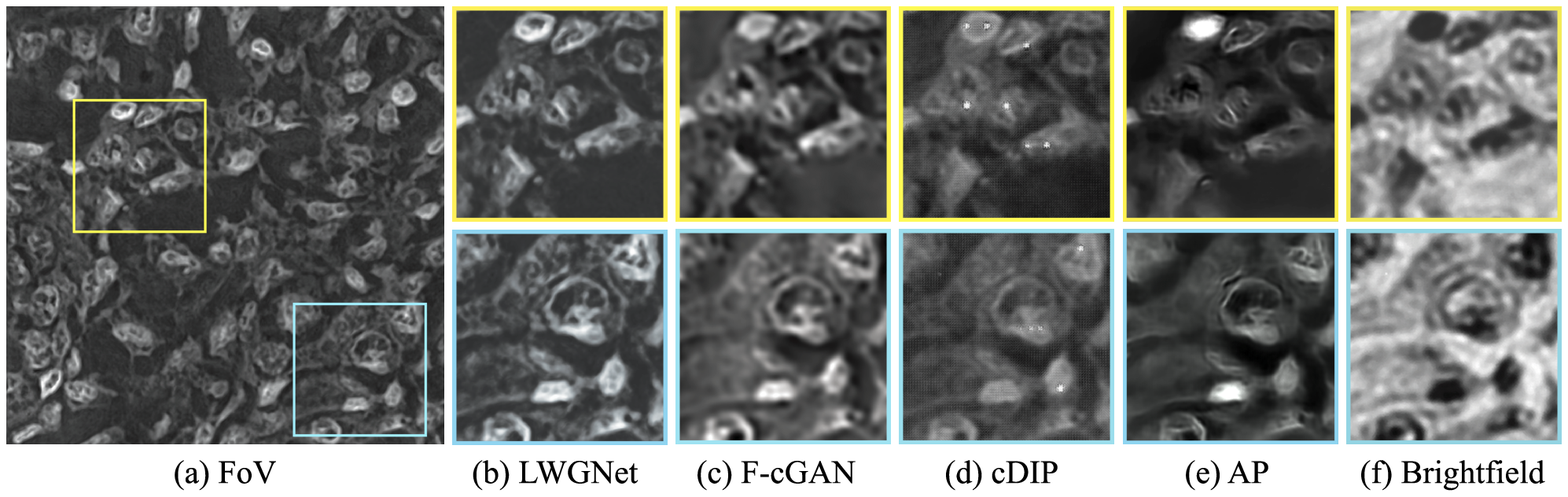}} \\
         
    \hline
    
    \end{tabular*} \\
    \vspace{2em}
    \captionof{figure}{Phase reconstruction from \textit{real} 16-bit measurements of Lung carcinoma, 12-bit Osteosarcoma, and 8-bit Osteosarcoma. (a) $1500\times1500$ pixels FoV reconstruction using proposed LWGNet, (b) zoomed reconstruction shown in inset, (c) reconstruction obtained from F-cGAN, (d) reconstruction obtained from cDIP, and (e) input brightfield image.}
    \label{fig:exp_results}
    \vspace{5em}
\end{table}

We test our model on real experimental data and compare the results against two deep-learning techniques, cDIP and F-cGAN. Prior to evaluation, we finetune all the methods on real dataset described in Section \ref{sec:real_dataset}. Figure \ref{fig:exp_results} show the visual results for 16, 12 and 8 bit depths respectively. For our experiments, we use histological sample slides, specifically Lung carcinoma and Osteosarcoma (H\&E stained).

Phase reconstruction obtained from cDIP show low contrast and are prone to checkerboard artifacts. F-cGAN reconstructs the cellular structures but performs relatively poor in reconstruction of finer background details. However, the proposed algorithm successfully preserves edges and background details compared to the other approaches consistently for all bit depths.
The computed gradients of peripheral LED images contain high-frequency details. We hypothesize that the learned complex-valued neural network parameters are optimized to map these useful gradients to desired object field more effectively than neural networks that use just the intensity images as input. Moreover, due to the extremely small parameter count of our proposed method, finetuning on a small dataset of real data doesn't lead to any overfitting.

\subsection{Sensor Quantization Analysis}
In this section, we experimentally verify that the proposed method shows little perceptual variation over reconstruction with increasing quantization noise or decrease in the bit depth of the input data. We finetune our proposed approach using the dataset described in Section \ref{sec:real_dataset} prior to evaluation. Fig \ref{fig:bit_depth_exp} shows reconstruction for Osteosarcoma (top) and Lung carcinoma (bottom) samples at different bit-depth settings of the camera. The reconstructions obtained are perceptually indistinguishable in the first case as shown in the inset. In the second case, we observe minor changes at reconstructions of sharp edge features with the bit-depth setting.
\begin{table}[!ht]
    \centering
    \begin{tabular*}{0.845\linewidth}{|c|}
    \hline
    \fcolorbox{white}{white}{ \includegraphics[width=0.8\textwidth]{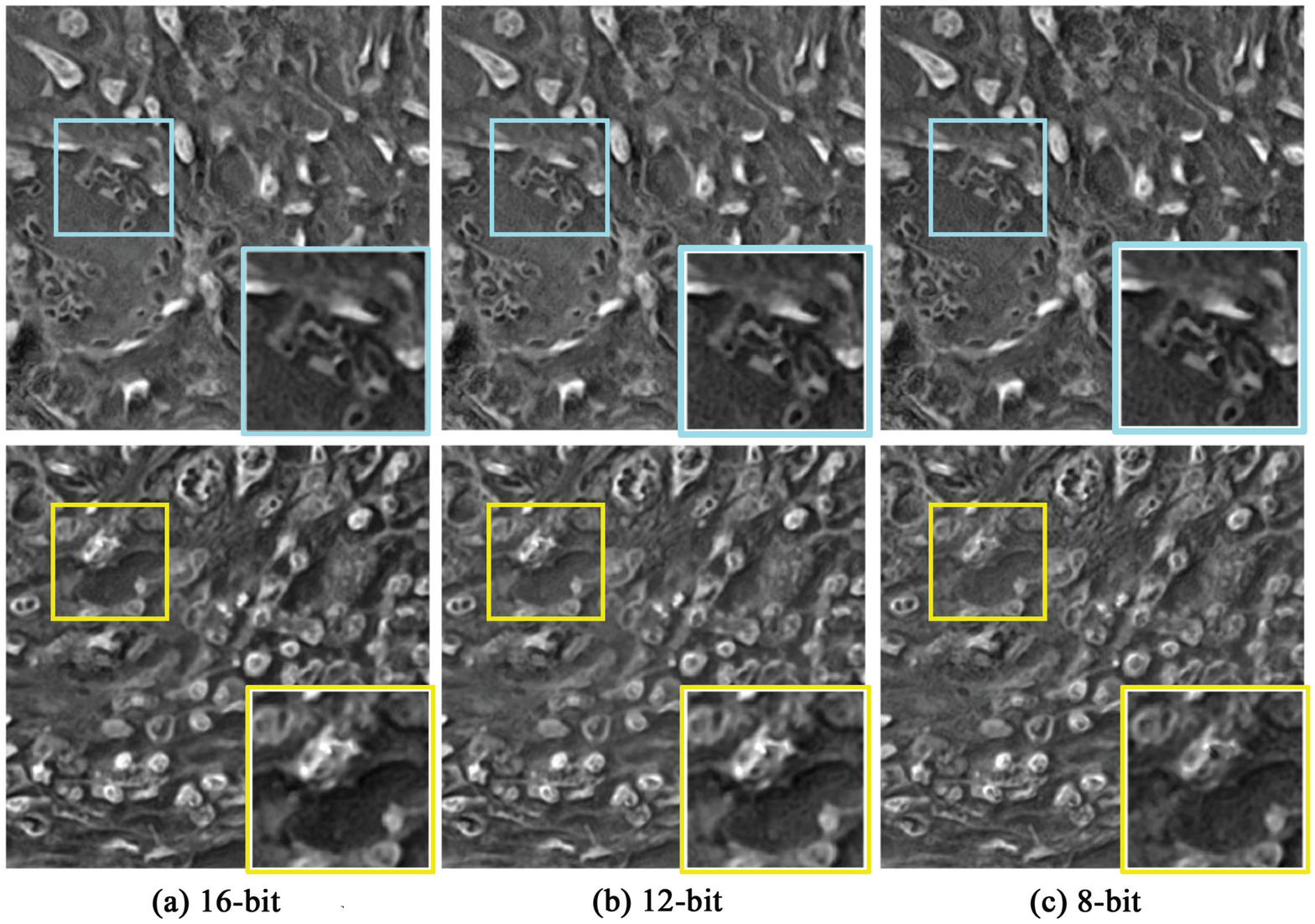}} \\
    \hline
    \end{tabular*} \\
    \vspace{2em}
    \captionof{figure}{Comparison of phase reconstruction using the proposed LWGNet on Osteosarcoma (top) and Lung carcinoma (bottom) at various bit depth settings. The reconstruction quality is consistent even under high quantization noise.}
    \label{fig:bit_depth_exp}
\end{table}

\newpage

\subsection{Ablation Study}
In this subsection, we analyse the impact of different components of our proposed network on simulated data.

\subsubsection{Effect of Complex-valued Operations}
\begin{figure}[!ht]
\begin{table}[H]
    \centering
    \begin{tabular}{|c|p{75mm}|p{75mm}|}
    \hline
     \hbox{\rotatebox{90}{\small Complex-network}} &\includegraphics[width=0.46\textwidth]{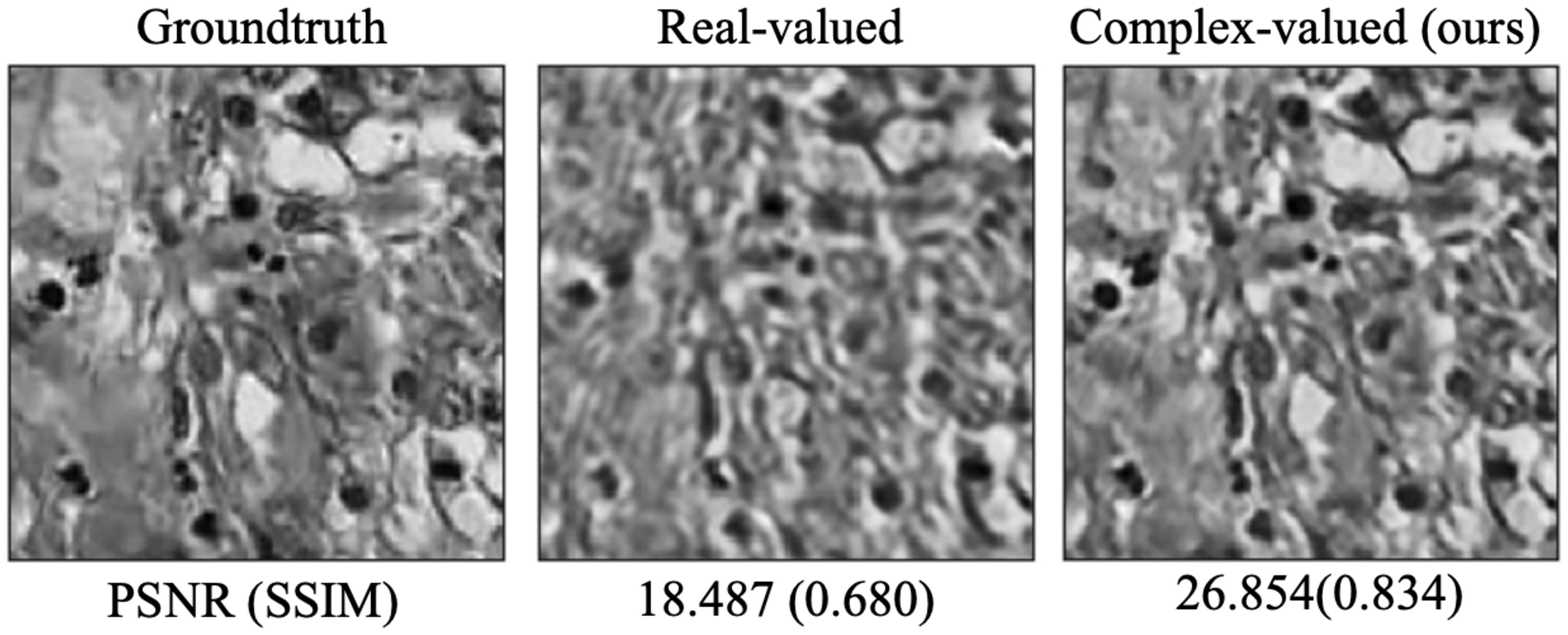} &
     \includegraphics[width=0.46\textwidth, trim={0.25 0 0 0}]{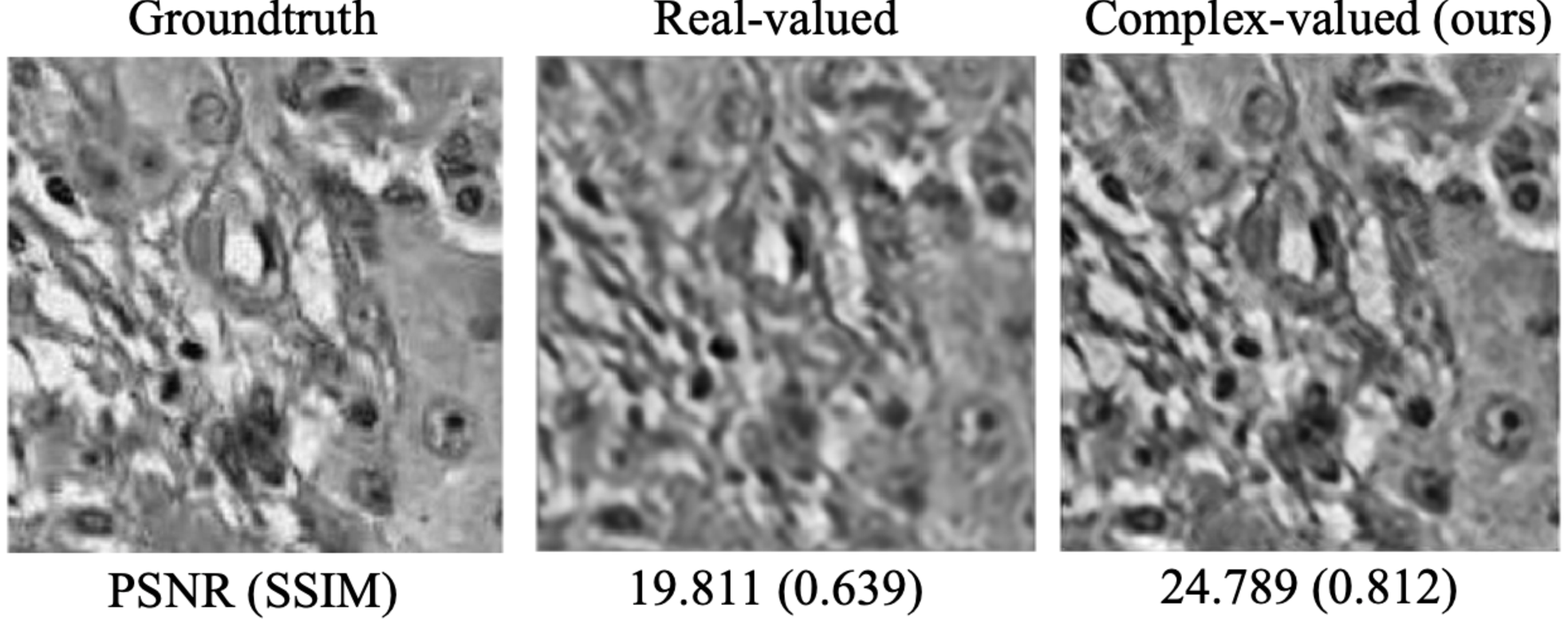}\\
     \hline
     \hbox{\rotatebox{90}{\small Gradient-processing}}
     &\includegraphics[width=0.47\textwidth]{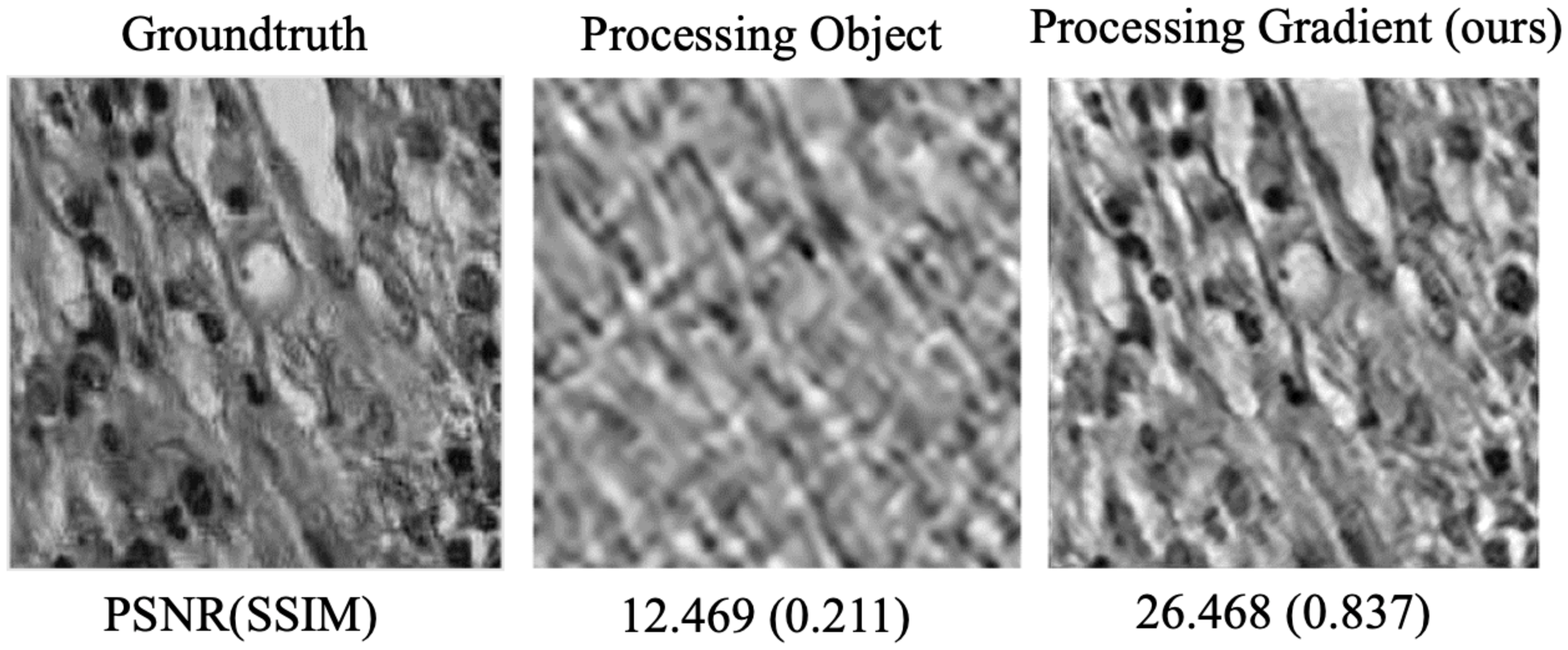} &
     \includegraphics[width=0.47\textwidth, trim={0.35cm 21.25cm 0.5cm 8.5cm}]{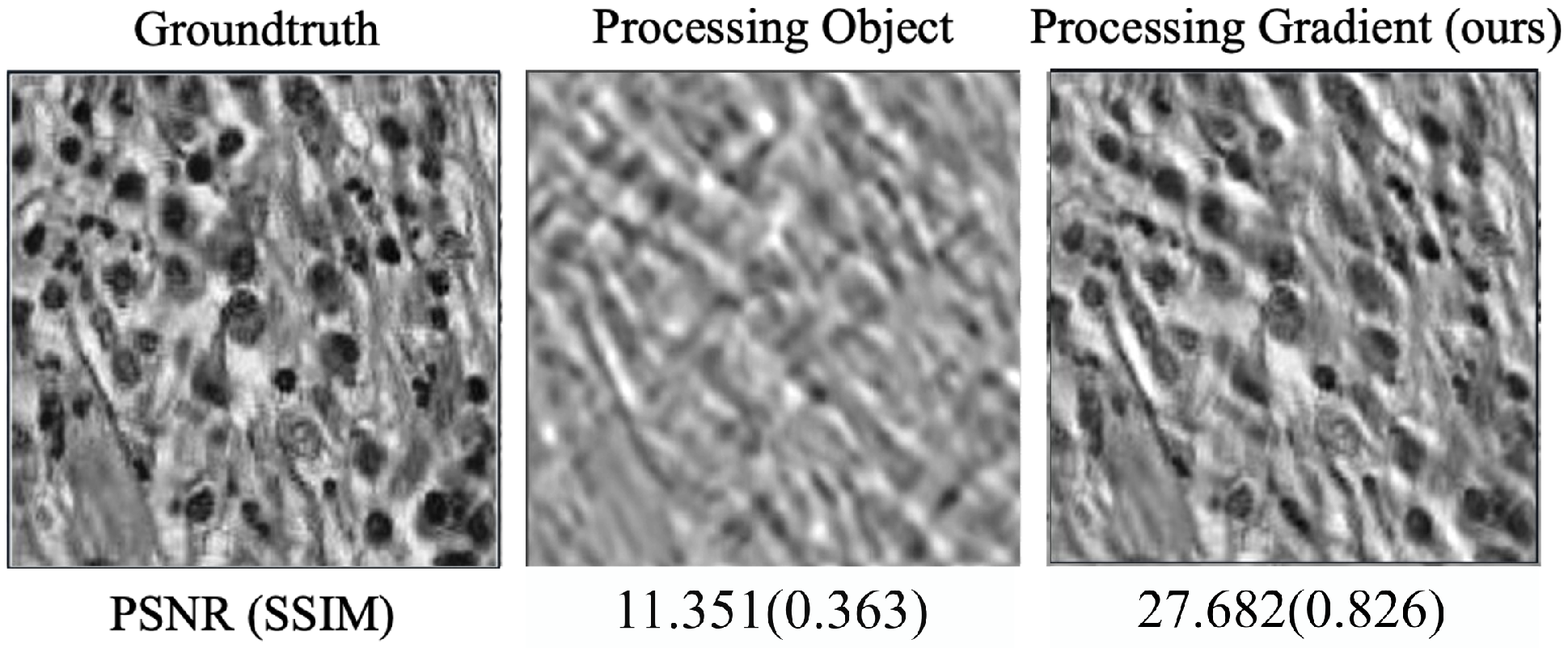}\\
     \hline
     \hbox{\rotatebox{90}{\small Stochastic gradient}}
     &\includegraphics[width=0.47\textwidth]{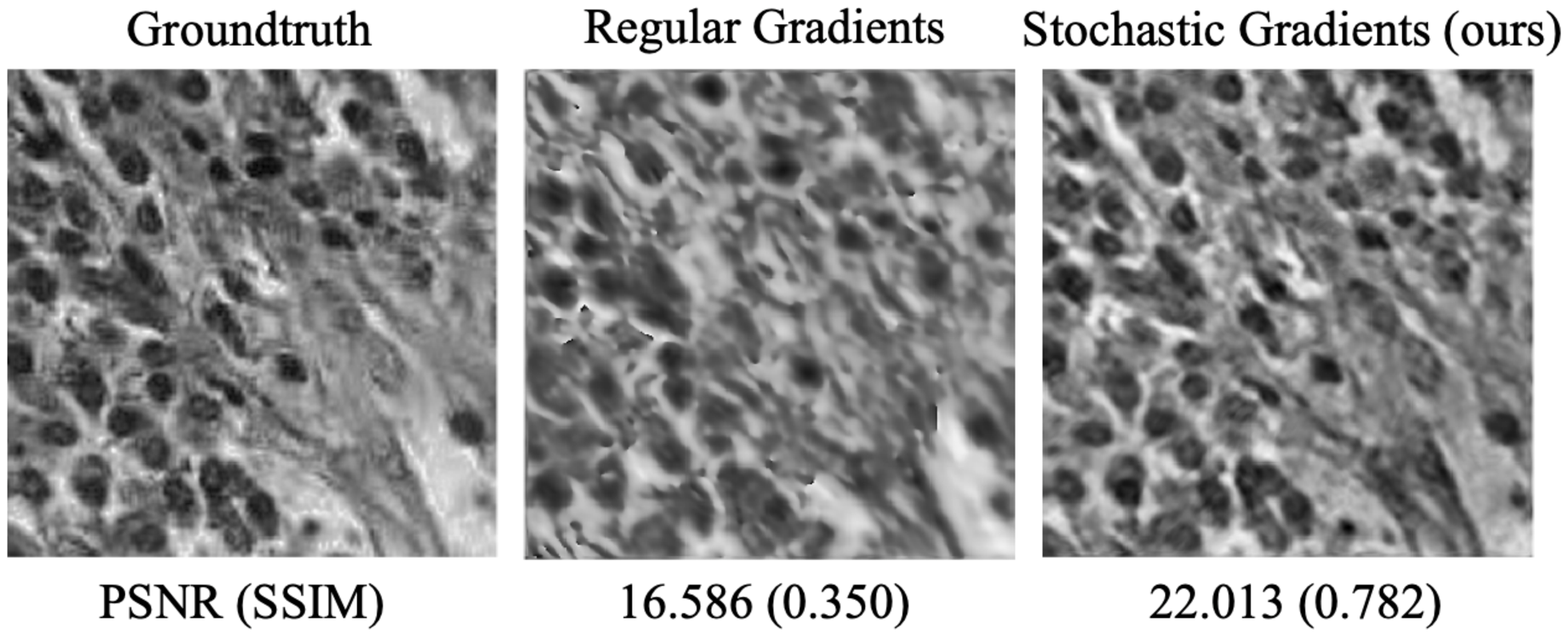} &
     \hspace{0.1em}\includegraphics[width=0.465\textwidth]{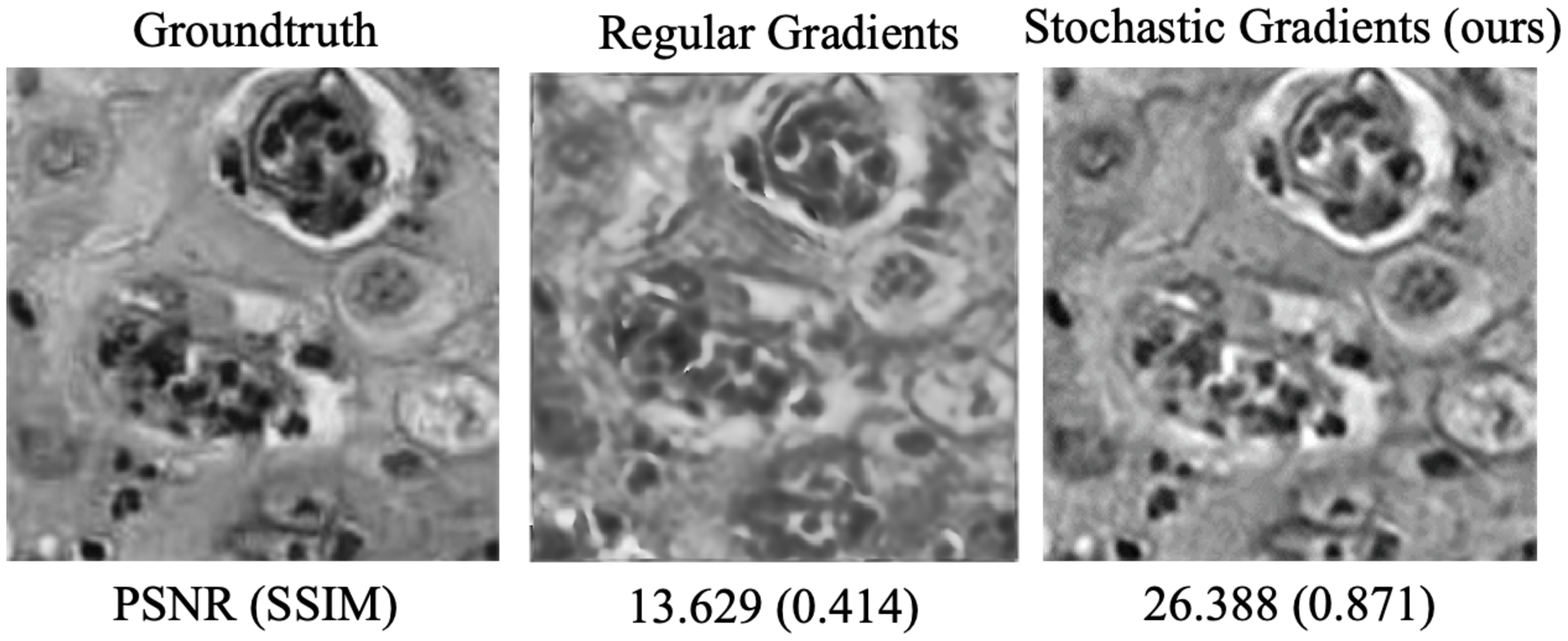}\\
     \hline
    \end{tabular}
    \vspace{2em}
    \captionof{figure}{Ablation experiments: First row shows the using complex-valued operations leads to improved reconstruction . Second row shows that the proposed method of learning the network on gradient leads to effective utilization of gradient information than learning a network on estimated object field. Third row shows that using stochastic gradient improves our results compared to regular gradients.}
    \label{fig:all_ablation}
\end{table}
\end{figure}

Here we verify the necessity of the complex-valued operations described in Section \ref{sec:method}. To do that, we replace the complex-valued operation of our proposed network with real-valued operations acting individually on the real and imaginary channels of the complex gradient. We keep the rest of the architecture and loss functions the same. Top row of Fig \ref{fig:all_ablation} presents the visual comparison between reconstructions obtained from complex-valued and real-valued networks. Our experiments show that the use of complex valued operations increases the average PSNR by ~5dB.

\subsubsection{Effect of Processing Gradients}
To verify that processing gradients through a learned network actually helps, we train a variant of the proposed architecture with the following update stage where the learned neural network acts on the estimated object field,
\begin{align}
    \mathcal{O}_{f}  = \hat{\Psi}(\mathcal{O}^{K})
    \label{eq:post_res_block}
\end{align}
where,
\begin{align}
\mathcal{O}^{i+1} = (\mathcal{O}^i + \frac{1}{L} \sum_{\mathcal{l}=1}^{L}\mathcal{g}_{\mathcal{l}}^i), i \in \{0,1,..,K-1\} \label{only_wf}
\end{align}
and $\hat{\Psi(.)}$ is $N$ numbers of $\Psi^{i}(.)$ (described in Section \ref{sec:method}) stacked together. We used $K = 5$ iterations of Wirtinger flow and stack $N=3$ numbers of $\Psi(.)$ for this purpose to compare against 3 stages of proposed approach. The rest of the architecture and loss functions are kept the same as proposed method. Second row in Fig \ref{fig:all_ablation} shows the visual results highlighting the effect of processing the gradients. The proposed way of learning a mapping from the stochastic gradients to object fields leads to more effective use of the gradient information compared to a enhancing an object field using a neural network after wirtinger updates.

\subsubsection{Effect of Stochastic Gradients}
In this experiment, we verify the efficacy of the stochastic gradients estimated in Eq.(\ref{eq:stoch_grad}). To do that, we train a variant of the proposed network where regular gradients are used instead of the stochastic gradients and each stage is given by,
\begin{align}
    \mathcal{O}^{i+1} = \mathcal{O}^i + \Psi^i([\nabla_{\mathcal{O}}f^{i}_{\mathcal{l}}]_{\mathcal{l}\in L}).\label{eq:av_res_block}
\end{align}
where $ \nabla_{\mathcal{O}}f^i_{\mathcal{l}}$ is given by Eq.(\ref{eq:grad_Wirtinger}).
The rest of the network and loss functions are kept the same as the proposed LWGNet. The third row of Fig \ref{fig:all_ablation} shows visual results for this experiment. The use of stochastic gradient clearly outperforms the use of regular gradients.
\subsubsection{Effect of Number of Unrolled Stages}
\begin{figure}[!ht]
\begin{table}[H]
    \centering
    \begin{tabular}{|c|c|}
    \hline
    \begin{tikzpicture}[scale=0.65]
    \begin{axis}[
    xlabel={Number of update stages},
    ylabel={PSNR},
    xmin=1, xmax=12,
    ymin=20, ymax=30,
    xtick={1, 2, 3, 4, 5, 6, 7, 8, 9, 10, 11},
    ytick={20, 22, 24, 26, 28, 30},
    legend pos=south east,
    ymajorgrids=true,
    xmajorgrids=true,
    grid style=dashed,
    ]
    \addplot[
        color=blue,
        mark=o,
        ]
        coordinates {
        (2,23.4753)(3,27.6838)(4,27.9985)(5,28.3477)(7,28.4082)(10,28.206)
        };
        \addlegendentry{16-bit}
    \addplot[
        color=red,
        mark=o,
        ] 
        coordinates {
        (2,23.4612)(3,27.3033)(4,28.0127)(5,28.2901)(7,28.3476)(10,28.1528)
        };
        \addlegendentry{12-bit}
    \addplot[
        color=violet,
        mark=o,
        ]
        coordinates {
        (2,21.1026)(3,22.8025)(4,25.7494)(5,27.0354)(7,27.8197)(10,27.952)
        };
        \addlegendentry{8-bit}
    \end{axis}
    \end{tikzpicture} &
    \fcolorbox{white}{white}{\includegraphics[width=0.435\textwidth]{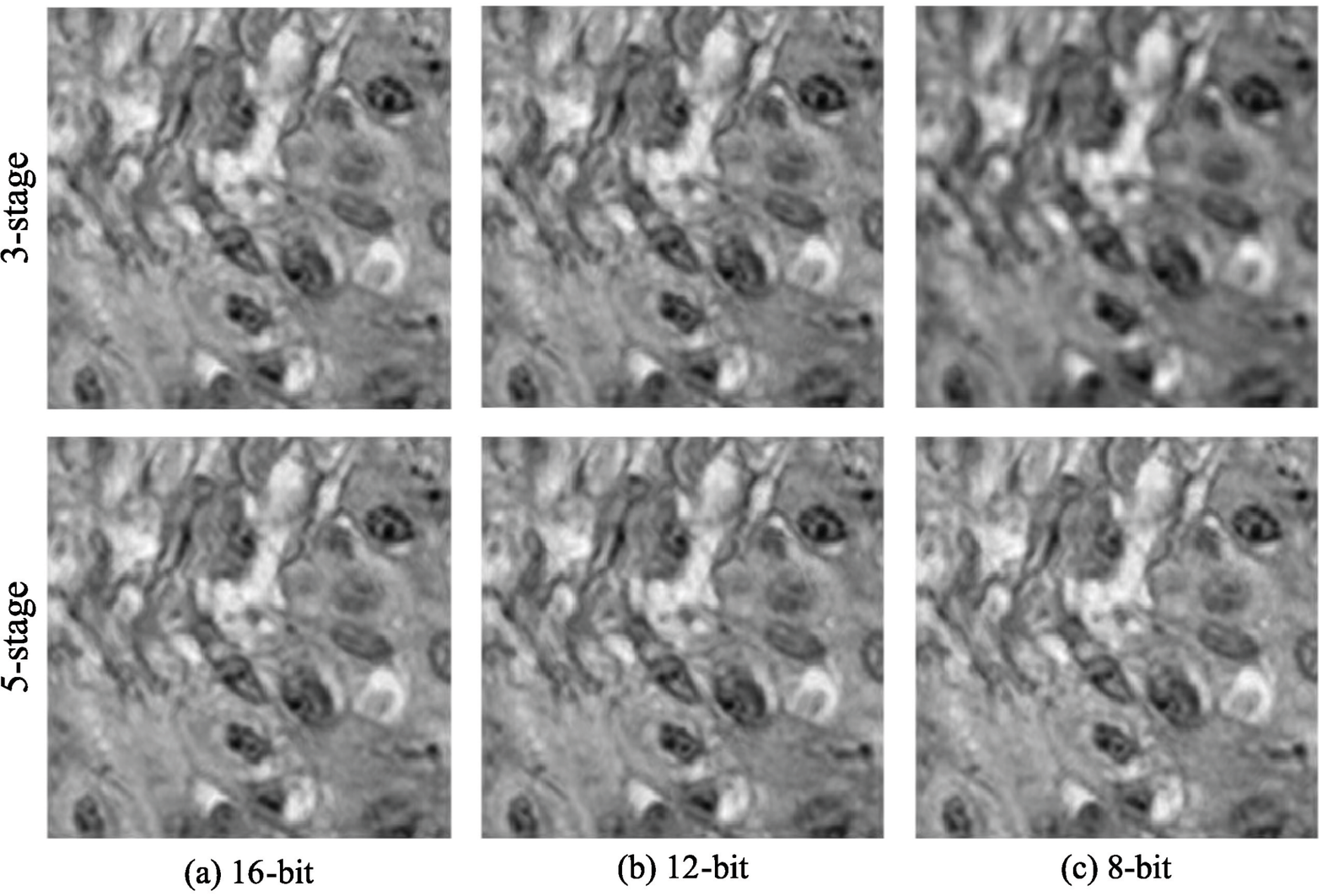}} \\
    \hline
    \end{tabular}
    \label{tbl:table_of_figures}
\end{table}
\caption{Left: PSNR vs Number of unrolled stages. Right: corresponding reconstructions for 3 and 5 stage LWGNet under different bit depths. For 16 and 12 bit data, after just 3 stages the performance saturates while 5 stages are needed for 8 bit data.}
\label{fig:stages_psnr}
\end{figure}
In this section, we analyse the effect of the number of unrolled iterative stages ($K$ in Section \ref{sec:method}). Fig \ref{fig:stages_psnr} represents the variation of PSNR with the number of update blocks in the proposed architecture varied between 2 to 10 blocks. We found that the PSNR saturates after 3 stages for 16-bit and 12-bit data, while it takes at least 5 stages to achieve a similar performance for 8 bit data. 
\section{Conclusion}
\label{sec: conclusion}
We propose a novel physics based neural network for FPM reconstructions. Our network derives inspiration from conventional wirtinger flow phase retrieval and combines it with data-driven neural networks. Unlike naive wirtinger flow, our network learns a non-linear mapping from stochastic gradients to object field intermediates. We use complex-valued neural networks to learn this non-linear mapping. We perform extensive experiments on simulated and real data to validate the proposed architecture's efficacy. Our method performs at par or even better than existing FPM reconstruction techniques especially for difficult scenarios like low dynamic range. We attribute the success of our network on such challenging conditions to the novel non-linear physics-based mapping. Moroever, we also collect a dataset of real samples using a low cost sensor under different sensor bit depths which will be made public upon acceptance of the paper. The ability to perform high-resolution wide-field of view microscopy using low-cost sensor through physics inspired data-driven techniques can significantly bring down the system cost and is a step towards making point of care diagnostics more accessible.

\section*{Acknowledgments}
We gratefully acknowledge the funding from DST IMPRINT-2 (IMP/2018/001168).

\bibliographystyle{unsrtnat}
\bibliography{references}  






\section{Our real dataset}
\label{sec: dataset_caps}

\begin{figure}[!ht]
    \centering
    \includegraphics[height=8cm]{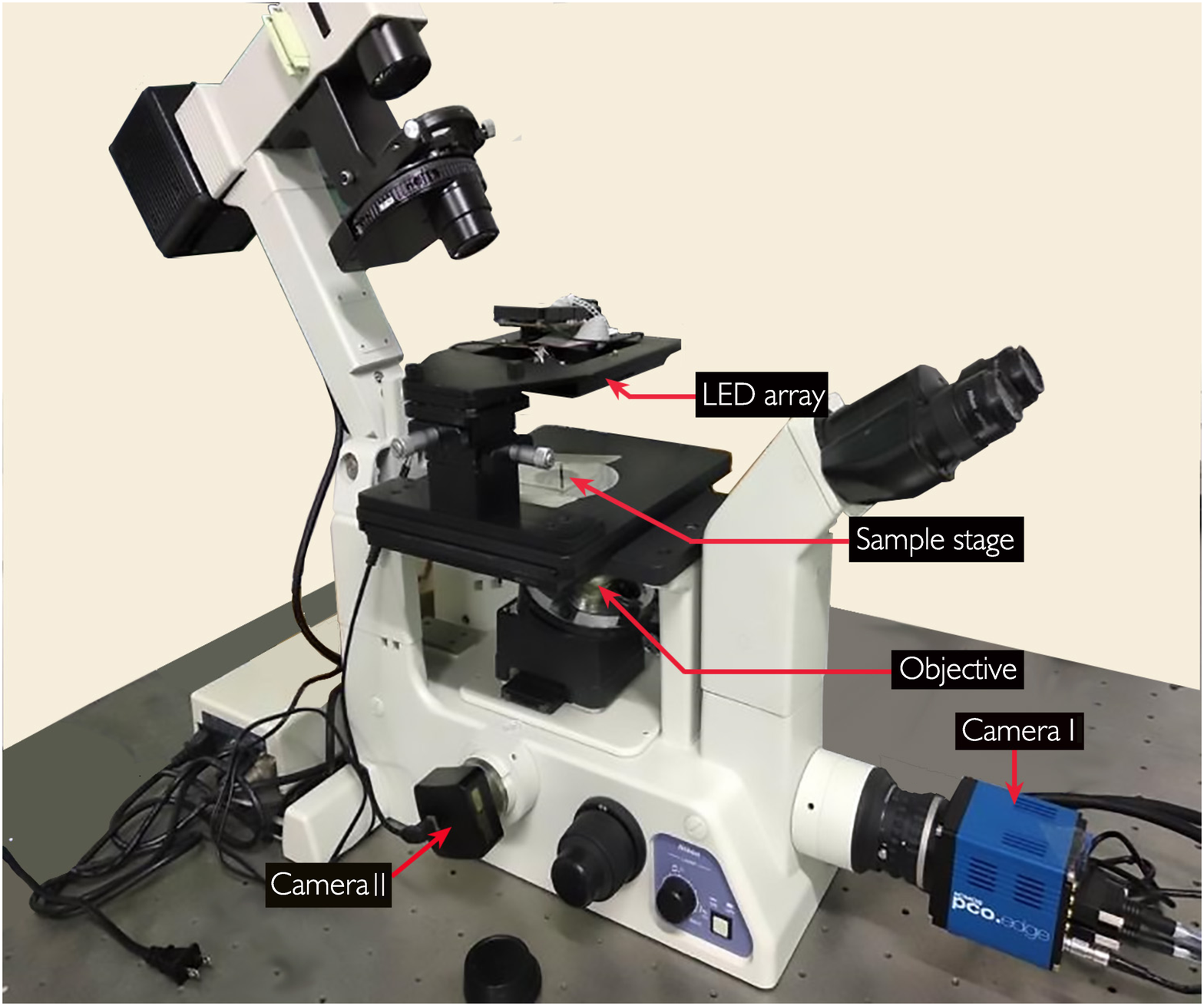}
    \caption{FPM Experimental setup. Camera I is the low cost low dynamic range machine vision camera to capture low bit-depth images while Camera II is the expensive high-dynamic range sCMOS camera that captures 16-bit images.}
    \label{fig:capture_setup}
\end{figure}

Fig \ref{fig:capture_setup} shows the experimental setup used for capturing the real dataset described in Section 4.2 of the main paper. Camera I is a low cost machine vision CMOS camera that allows 3 bit depth settings - 8, 12, and 16. Camera II is high cost sCMOS camera that only allows 16-bit imaging.

Fig \ref{fig:train_slides} and \ref{fig:test_slides} shows some histological samples used in our real data experiments. Here, we only show the brightfield images and the corresponding phase reconstructions for the sCMOS camera.

For obtaining the real data pairs used for finetuning and testing the proposed approach, we used 2 slides each of lung carcinoma, osteosarcoma, and cervical cells, and 1 slide of cerebral cortex. 1 slide each of lung carcinoma and cerebral cortex and 2 slides of cervical cells were used for training and the rest were kept for testing. From each slide, multiple regions were captured by translating the slides laterally. Specifically, from the training slides, 4 regions were captured while from the test slides 5 different regions were captured. For each region, 225 low-resolution sequential images were obtained corresponding to the 225 LEDs using 4 different imaging settings: 3 using the 3 bit-depth (8,12 and 16-bit) settings of camera I and 1 using 16-bit setting of camera II. The low resolution images from the two camera were registered to account for the misalignment. 

For obtaining groundtruth, AP reconstruction\cite{Tian:14} was performed on the sequential FPM measurements captured using the \textit{sCMOS} camera. Finally, before training, the 4 regions were divided into 640 non-overlapping patches. During training on 16-bit captured images the entire FoV of Camera I of size $1552\times2080$ pixels was considered. The entire FoV was divided into 160 samples, each of size $128\times128$ pixels. 
For fine-tuning the model on lower bit-depth (12-bit and 8-bit) images, the same slides were reused.
For inference on the 5 different regions of size $688\times688$ pixels FoV were cropped from the 3 slides shown in Fig \ref{fig:test_slides}. Each of these were again divided into 64 patches of size $128\times128$ with overlap of 48 pixels.
In Fig \ref{fig:test_slides}, exemplar FoV representing each of 3 slides is shown.
\begin{table}[!ht]
    \centering
    \begin{tabular*}{0.95\linewidth}{|c|c|c|c|c|}
    \hline
         \hbox{\rotatebox{90}{\hspace{1em}LED0 image}} &  
         \includegraphics[width=0.2\linewidth]{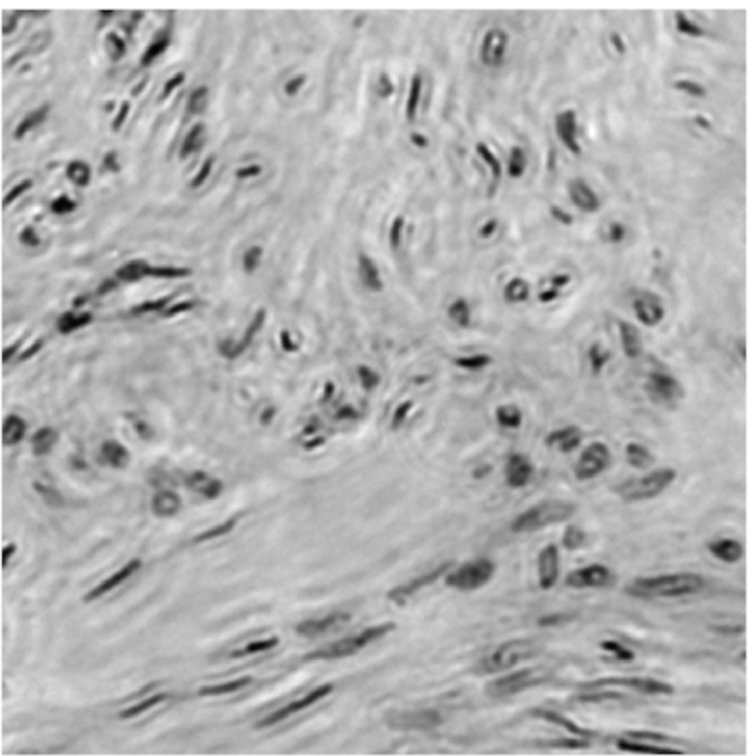} &
         \includegraphics[width=0.2\linewidth]{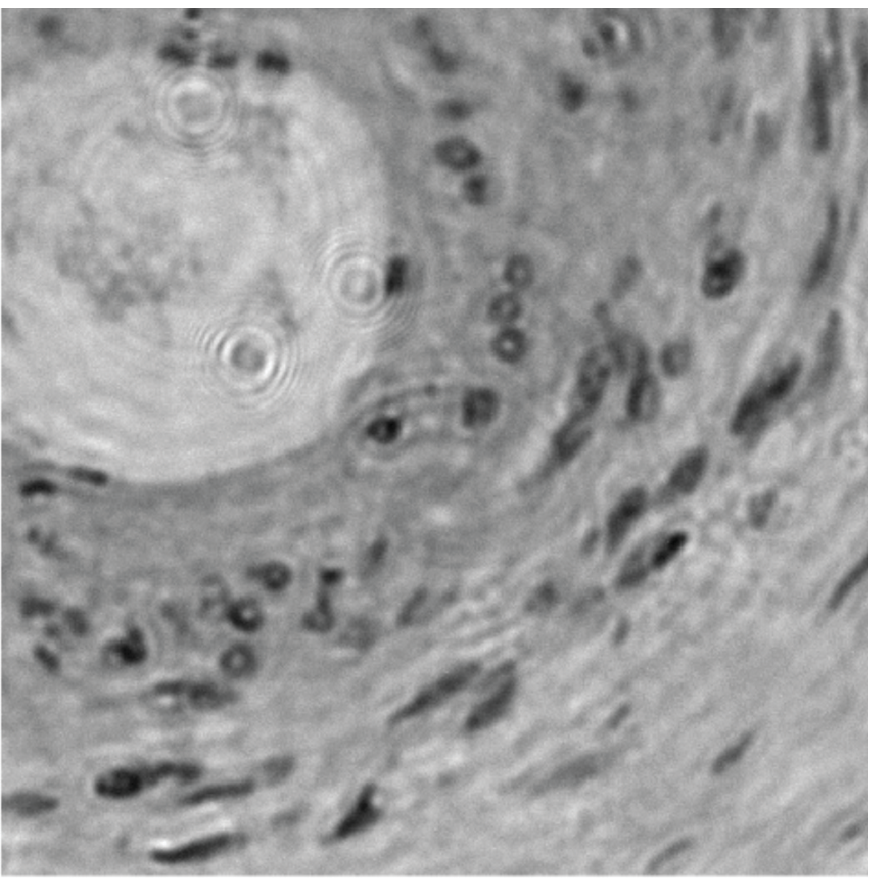} &  
         \includegraphics[width=0.2\linewidth]{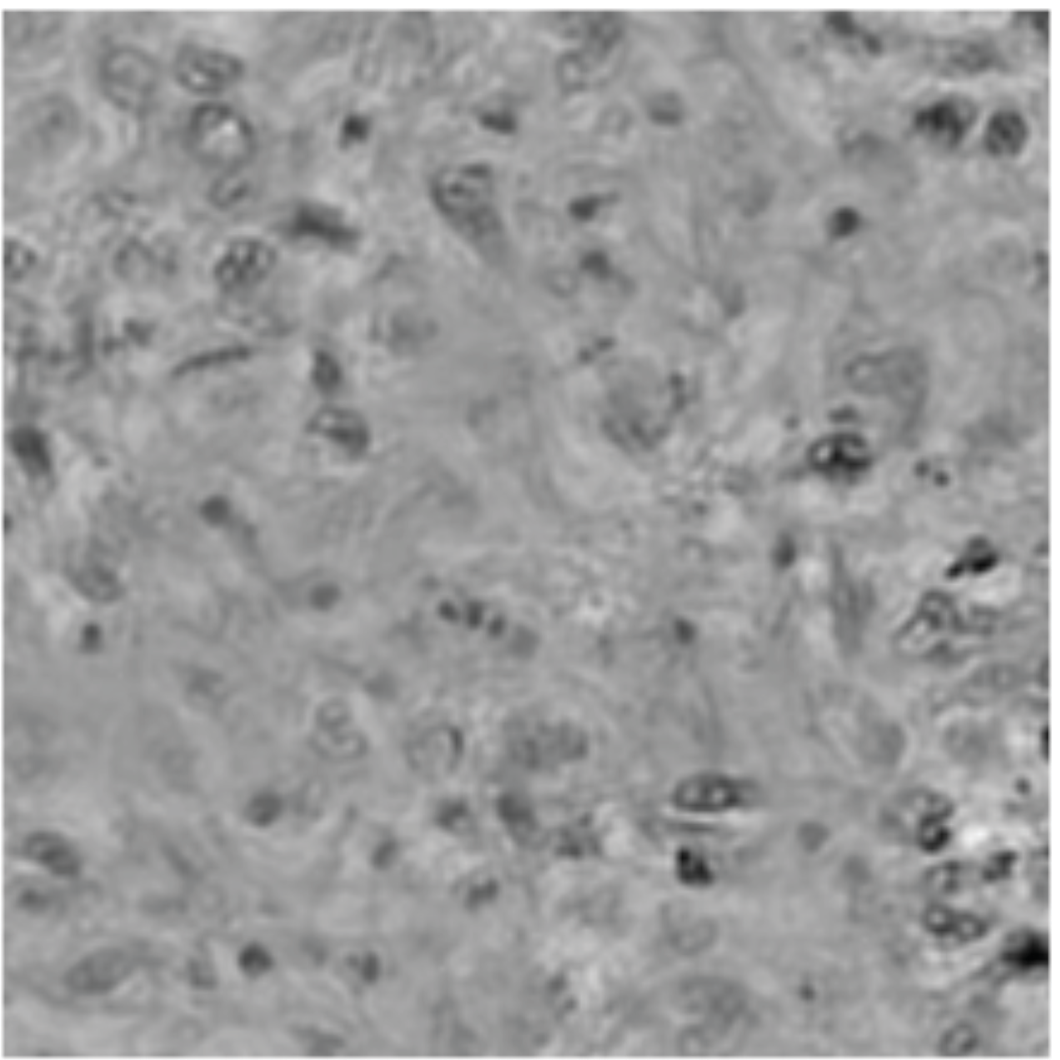} &
         \includegraphics[width=0.2\linewidth]{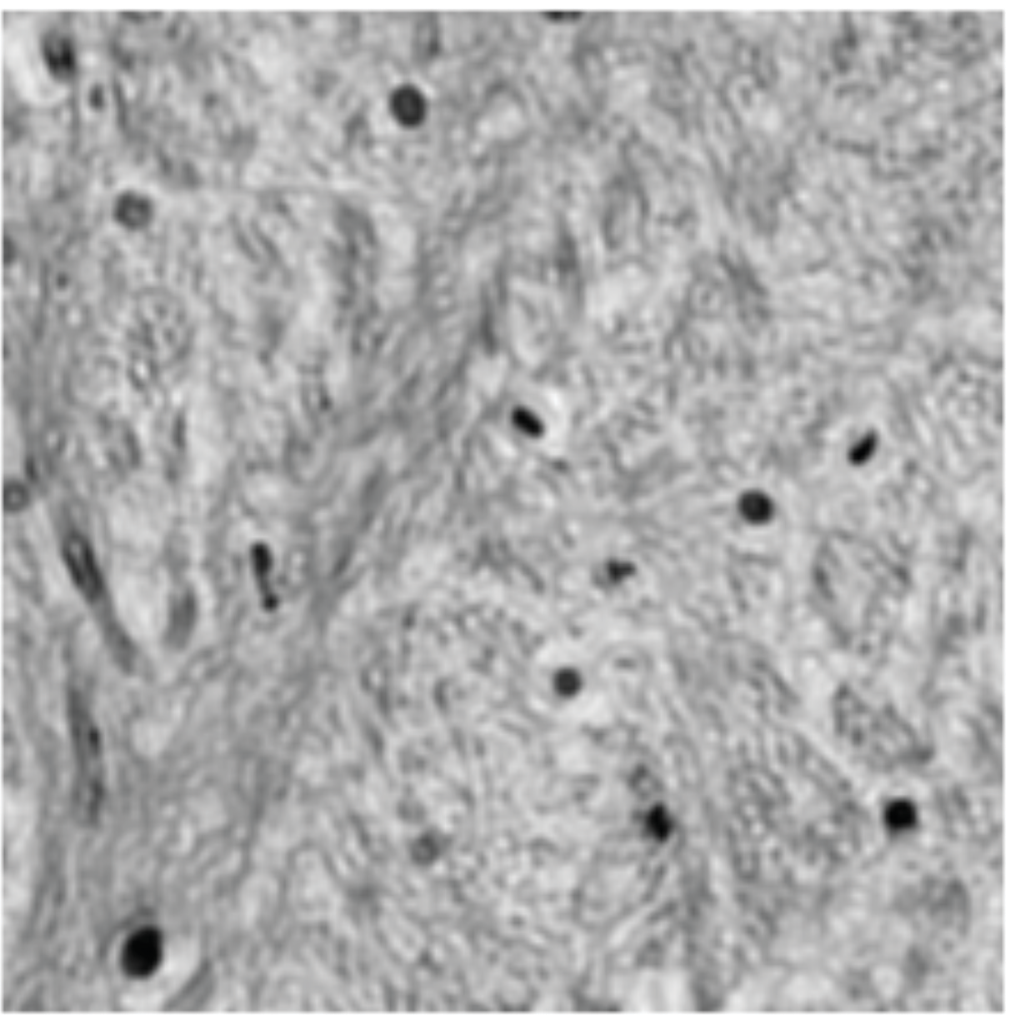}
         \\
    \hline
         \hbox{\rotatebox{90}{\hspace{1em}Phase}} &
         \includegraphics[width=0.2\linewidth]{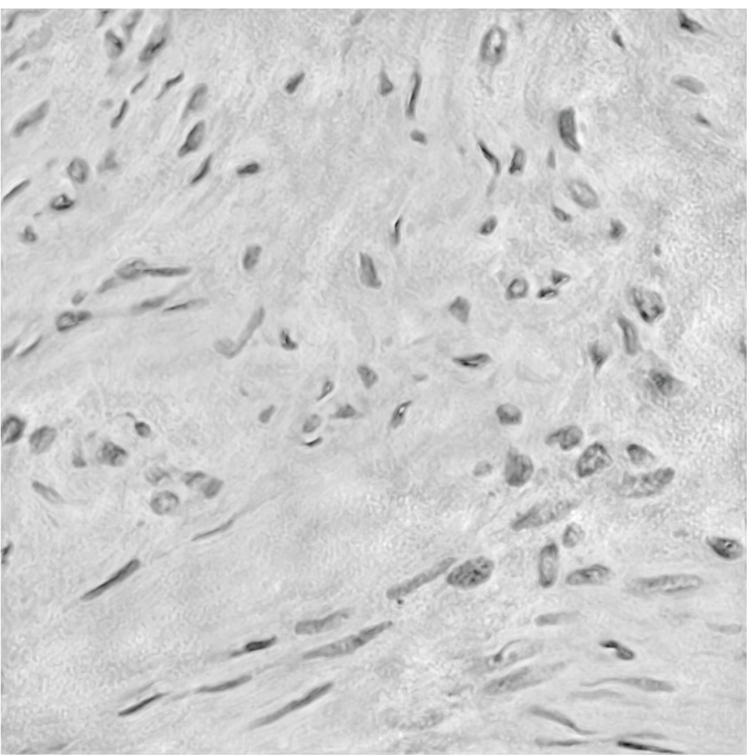} &
         \includegraphics[width=0.2\linewidth]{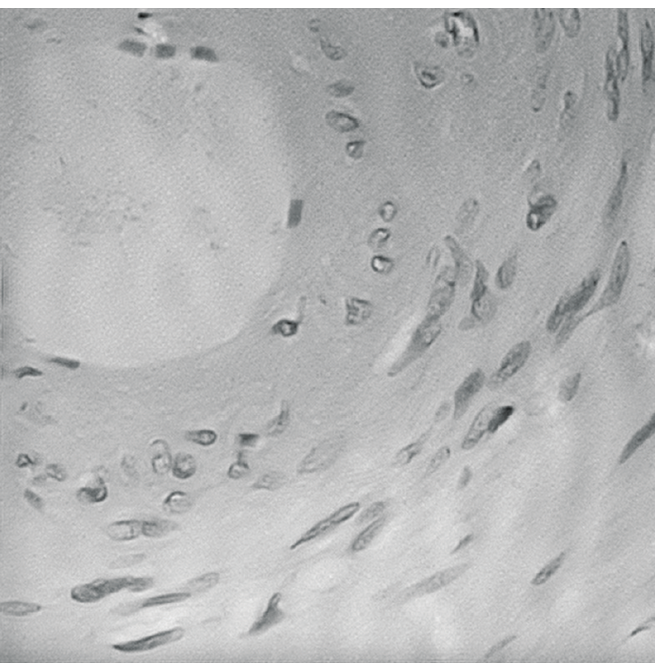} &
         \includegraphics[width=0.2\linewidth]{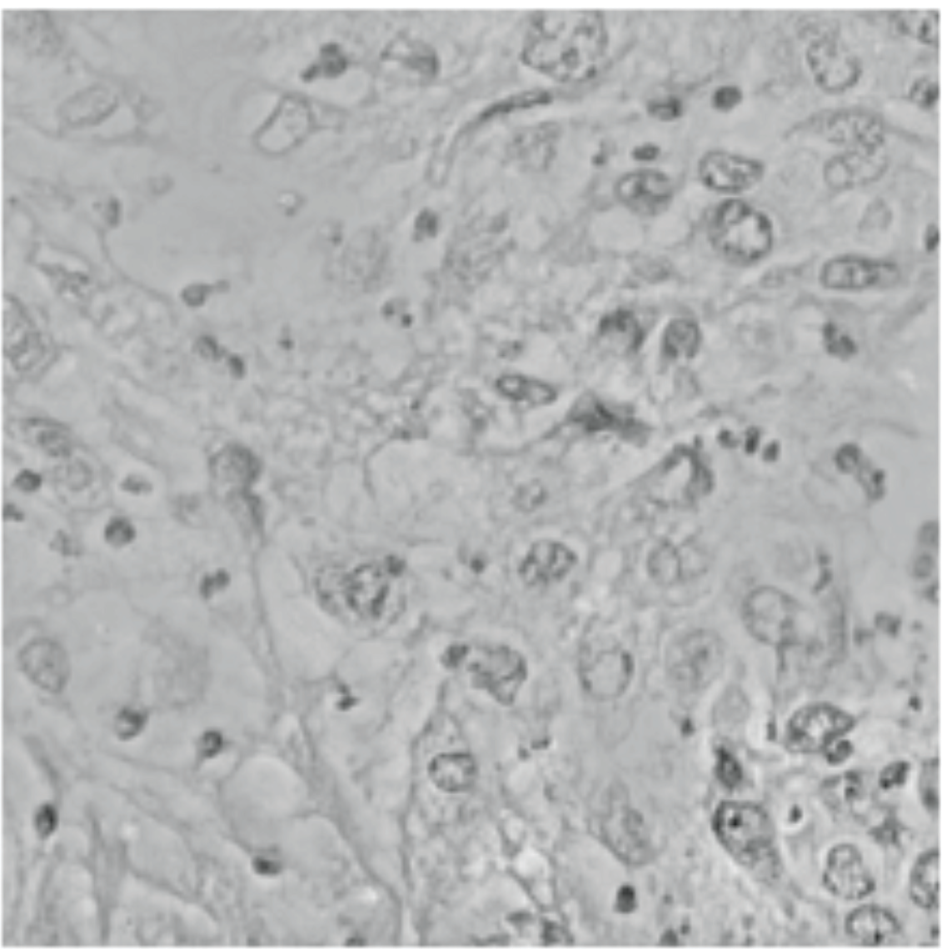} &
         \includegraphics[width=0.2\linewidth]{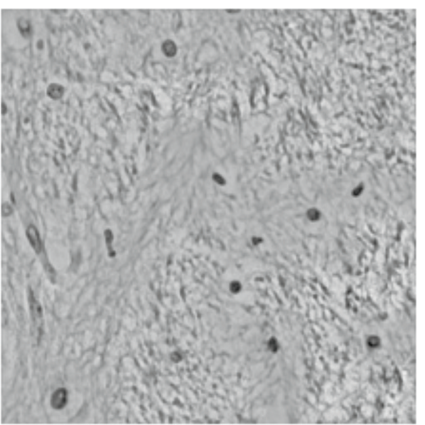}

         \\
         
         & \scriptsize{Cervical-Cells-1} & \scriptsize{Cervical-Cells-2} & \scriptsize{Lung-Carcinoma-1} & \scriptsize{Cerebral-Cortex-1}\\
    \hline
    
    \end{tabular*}
    \captionof{figure}{Histological samples used for training. Top row shows the brightfield images captured using sCMOS camera while the bottom row shows the corresponding AP phase reconstructions.}
    \label{fig:train_slides}
\end{table}

\begin{table}[H]
    \centering
    \begin{tabular*}{0.875\linewidth}{|c|c|c|c|}
    \hline
         \hbox{\rotatebox{90}{\hspace{1em}LED0 image}} &  \includegraphics[width=0.25\linewidth]{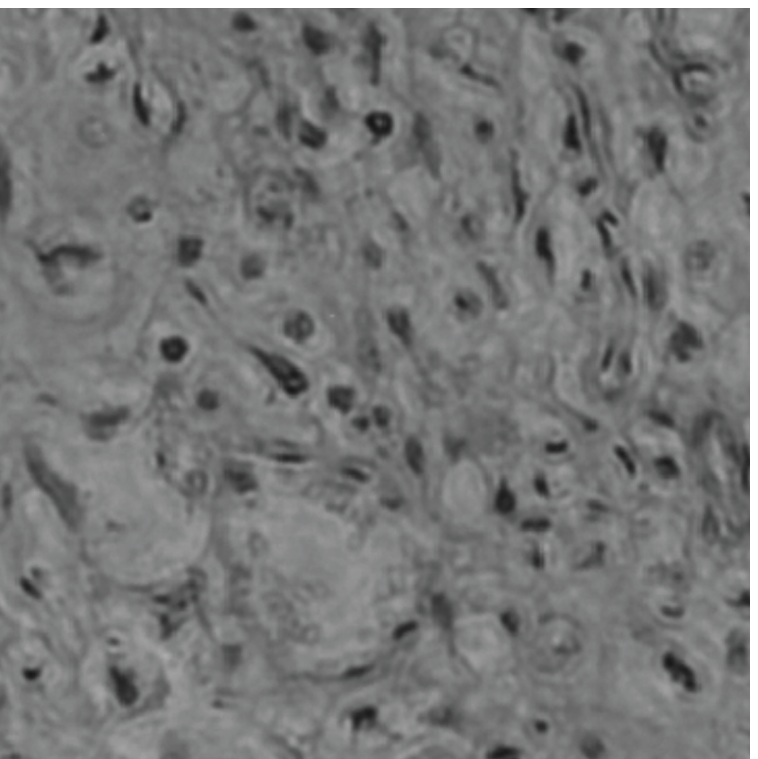} &
         \includegraphics[width=0.25\linewidth]{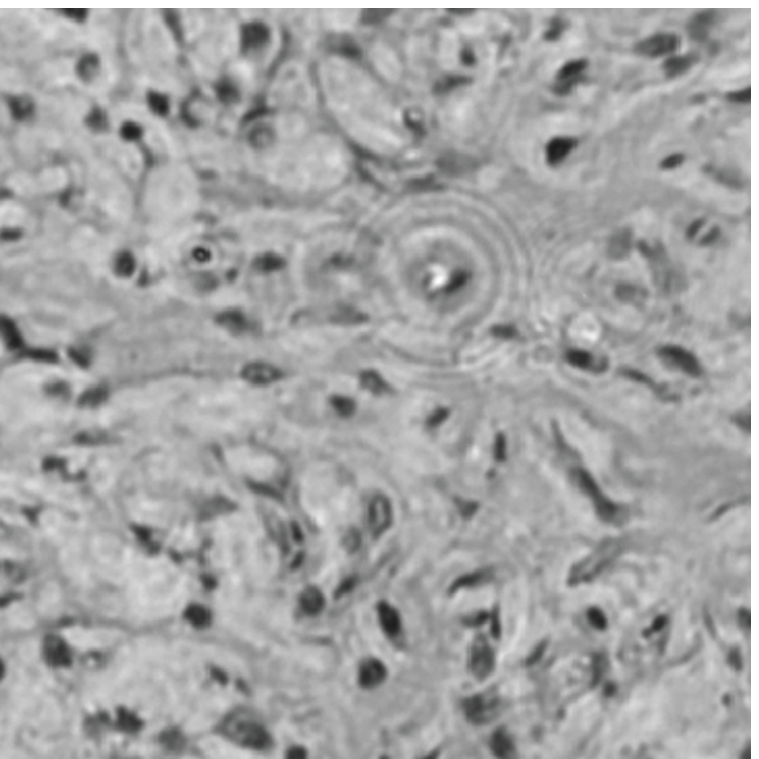} &
         \includegraphics[width=0.25\linewidth]{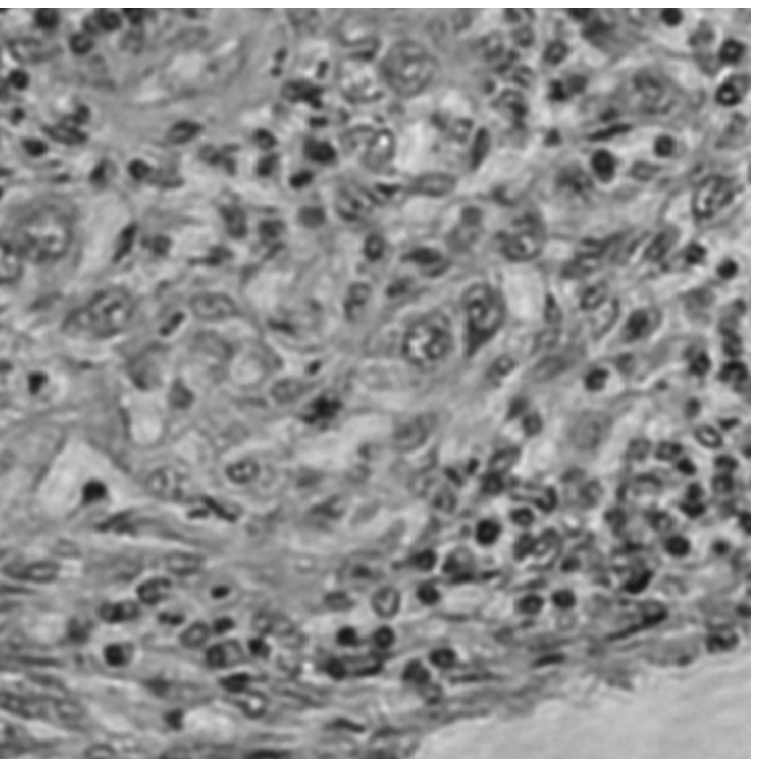}
         \\
    \hline
         \hbox{\rotatebox{90}{\hspace{1em}Phase}} &
         \includegraphics[width=0.25\linewidth]{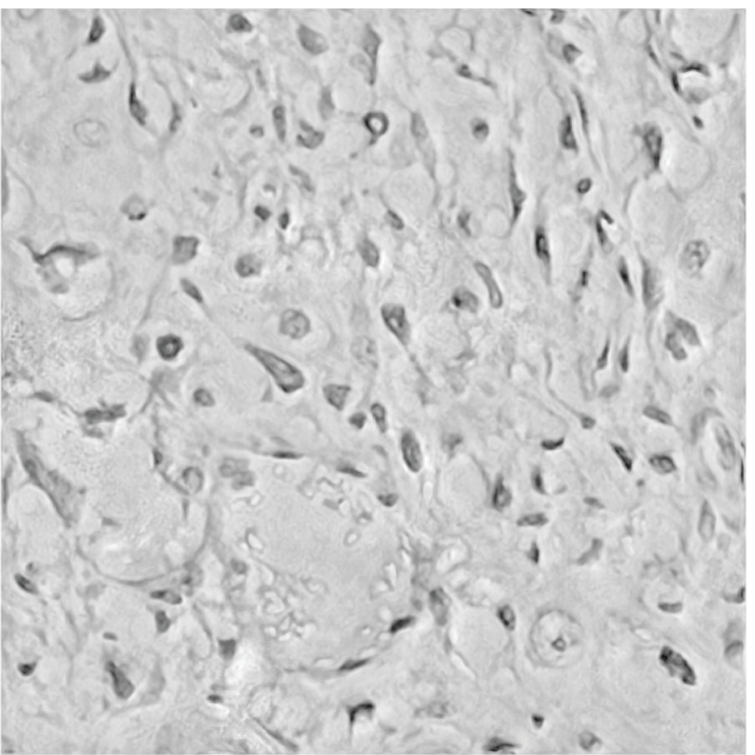} &
         \includegraphics[width=0.25\linewidth]{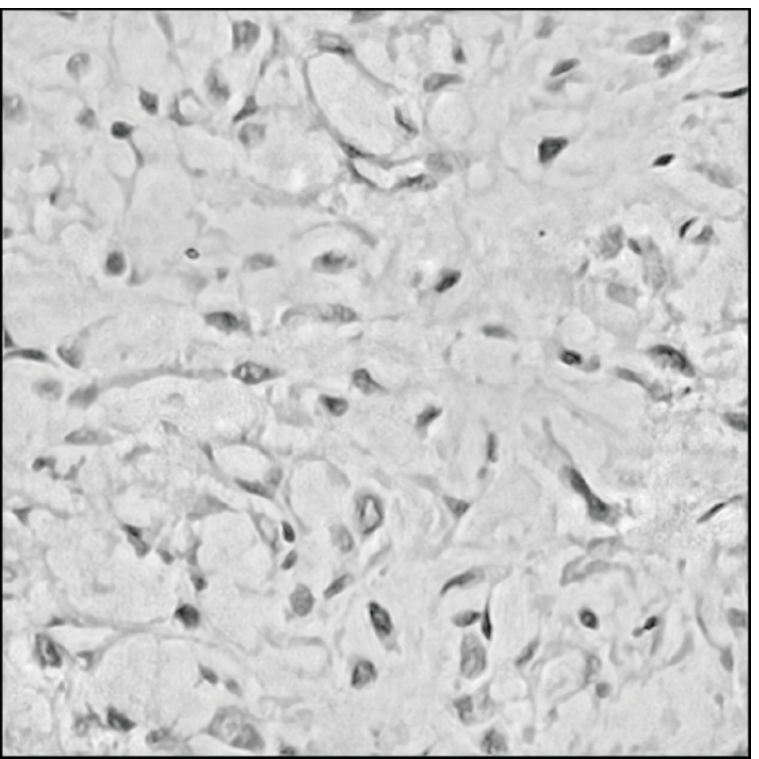} &
         \includegraphics[width=0.25\linewidth]{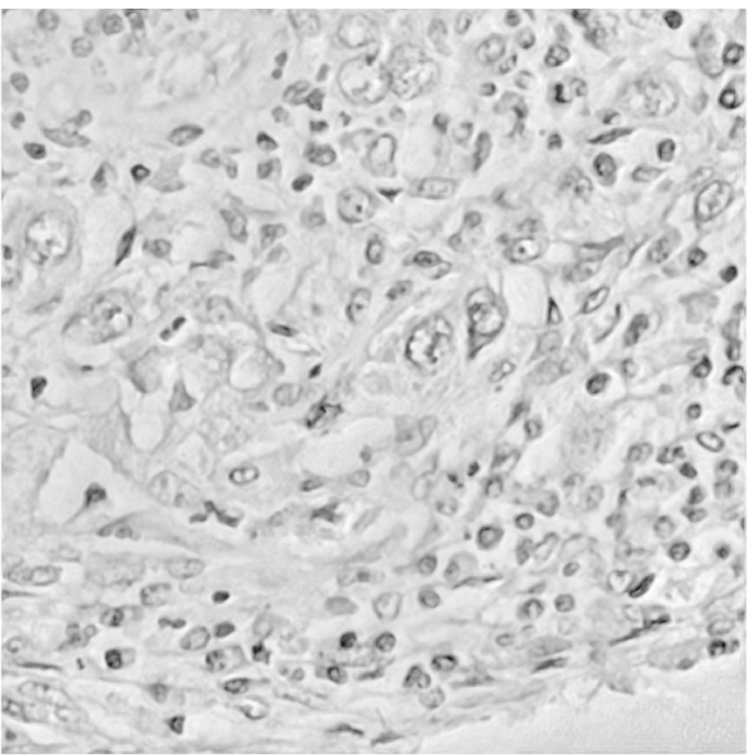} 
         \\
         
         & \scriptsize{Osteosarcoma-1} & \scriptsize{Osteosarcoma-2} & \scriptsize{Lung-Carcinoma-2} \\
    \hline
    
    \end{tabular*}
    
    \captionof{figure}{Histological samples used for testing. Top row shows the brightfield images captured using sCMOS camera while the bottom row shows the corresponding AP phase reconstructions.}
    \label{fig:test_slides}
\end{table}

\end{document}